\documentclass[sigconf]{acmart}

\AtBeginDocument{%
  }

\usepackage{amsthm}
\newtheorem{theorem}{Theorem}[section]

\newtheorem{proposition}[]{Proposition}[section]

\newtheorem{definition}{Definition}[section]

\usepackage{booktabs}
\usepackage{graphicx}
\usepackage{adjustbox}
\usepackage{amsmath}
\usepackage{bm}
\usepackage{color} 
\usepackage{tablefootnote}

\usepackage[ruled, vlined, linesnumbered]{algorithm2e}
\makeatletter 
\g@addto@macro{\@algocf@init}{\SetKwInOut{Parameter}{Learnable Parameters}} 
\makeatother
\usepackage{comment}
\usepackage{subcaption}
\usepackage{algorithmic}
\usepackage{enumitem}
\usepackage{colortbl}
\usepackage{float}
\usepackage{stfloats}

\newcommand{\cemp}[1]{{\cellcolor{blue!30}#1}}
\copyrightyear{2024}
\acmYear{2024}
\setcopyright{acmlicensed}\acmConference[KDD '24]{Proceedings of the 30th ACM SIGKDD Conference on Knowledge Discovery and Data Mining}{August 25--29, 2024}{Barcelona, Spain}
\acmBooktitle{Proceedings of the 30th ACM SIGKDD Conference on Knowledge Discovery and Data Mining (KDD '24), August 25--29, 2024, Barcelona, Spain}
\acmDOI{10.1145/3637528.3671849}
\acmISBN{979-8-4007-0490-1/24/08}

\begin{document}

\title{PolyFormer: Scalable Node-wise Filters via Polynomial Graph Transformer}
\subtitle{[Technical Report]}


\author{Jiahong Ma}
\affiliation{%
  \institution{Renmin University of China}
  \city{Beijing}
  \country{China}}
\email{jiahong_ma@ruc.edu.cn}

\author{Mingguo He}
\affiliation{%
  \institution{Renmin University of China}
  \city{Beijing}
  \country{China}}
\email{mingguo@ruc.edu.cn}

\author{Zhewei Wei}
\affiliation{%
  \institution{Renmin University of China}
  \city{Beijing}
  \country{China}}
\authornote{Zhewei Wei is the corresponding author. The work was partially done at Gaoling School of Artificial Intelligence, Beijing Key Laboratory of Big Data Management and Analysis Methods, MOE Key Lab of Data Engineering and Knowledge Engineering, and Pazhou Laboratory (Huangpu), Guangzhou, Guangdong 510555, China.}
\email{zhewei@ruc.edu.cn}


% \begin{abstract}
%   A clear and well-documented \LaTeX\ document is presented as an
%   article formatted for publication by ACM in a conference proceedings
%   or journal publication. Based on the ``acmart'' document class, this
%   article presents and explains many of the common variations, as well
%   as many of the formatting elements an author may use in the
%   preparation of the documentation of their work.
% \end{abstract}

\begin{abstract}
Spectral Graph Neural Networks have demonstrated superior performance in graph representation learning. However, many current methods focus on employing shared polynomial coefficients for all nodes, i.e., learning \textit{node-unified filters}, which limits the filters' flexibility for node-level tasks. The recent DSF attempts to overcome this limitation by learning node-wise coefficients based on positional encoding. However, the initialization and updating process of the positional encoding are burdensome, hindering scalability on large-scale graphs. In this work, we propose a scalable \textit{node-wise filter}, \textbf{PolyAttn}. Leveraging the attention mechanism, PolyAttn can directly learn node-wise filters in an efficient manner, offering powerful representation capabilities. Building on PolyAttn, we introduce the whole model, named \textbf{PolyFormer}. In the lens of Graph Transformer models, PolyFormer, which calculates attention scores within nodes, shows great scalability. Moreover, the model captures spectral information, enhancing expressiveness while maintaining efficiency. With these advantages, PolyFormer offers a desirable balance between scalability and expressiveness for node-level tasks. Extensive experiments demonstrate that our proposed methods excel at learning arbitrary node-wise filters, showing superior performance on both homophilic and heterophilic graphs, and handling graphs containing up to 100 million nodes. The code is available at \url{https://github.com/air029/PolyFormer}.
\end{abstract}



\maketitle

\section{Introduction}

% -----------------------------------KDD version 7-----------------------------------------
% paragraph 1
Graph Neural Networks (GNNs) have emerged as a powerful tool for addressing a variety of graph-related tasks, including node classification \citep{gcn_kipf2016semi, gat, graphsage}, link prediction \citep{seal_zhang2018link, dynamicgraph_FCS}, and graph classification \citep{gin}. GNNs are generally classified into two categories: spatial-based and spectral-based \citep{bernnet}. Spatial-based GNNs utilize message passing in the spatial domain for information aggregation, while spectral-based GNNs employ graph filtering operations in the spectral domain. Recently, Graph Transformer, a variant of the Transformer architecture \citep{transformer} adapted for graph data following its success in natural language processing \citep{bert, gpt3}, computer vision \citep{vit, swintransformer_liu2021swin}, and audio processing \citep{speechtransformer_dong2018speech, conformer}, has demonstrated superior performance in graph representation learning in both spatial and spectral manners \citep{san, graphormer, graphit_mialon2021graphit, gps_rampavsek2022recipe}.

% paragraph 2
Spectral GNNs have shown improved performance across various graph tasks, especially at the node level \citep{chien2021GPR-GNN, bernnet, chebnetii_he2022convolutional, optbasis_guo2023graph}. These approaches utilize different polynomial bases to approximate graph convolution and perform graph filtering on raw node signals. Despite their advancements, most models adopt shared polynomial coefficients for all nodes, leading to \textit{node-unified filters}. Alternatively, learning specific filters for each node, or \textit{node-wise filters}, offers greater flexibility \citep{GSP_isufi2022graph}. Though efforts of employing different filters for different signal channels \citep{jacobi, optbasis_guo2023graph} have been made, limitations with node-unified filters persist. A naive approach to learning node-wise filters involves excessive parameters, such as different polynomial coefficients for each node, leading to scalability issues. Besides, the learning process lacks generalization across nodes. To address these challenges, DSF \citep{dsf_guo2023graph} has attempted to employ a shared network on positional encoding, i.e., random walk positional encoding or Laplacian eigenvectors encoding, to learn node-wise polynomial coefficients, demonstrating the superiority of node-wise filters over node-unified filters. Nonetheless, the positional encoding update process is computationally intensive, with a complexity of $O(N^2)$, and its initialization, like obtaining Laplacian eigenvectors, also costs expensively with a complexity of $O(N^3)$, where $N$ represents the number of nodes. These hinder DSF's scalability for large graphs. Moreover, the performance of DSF heavily depends on the initial position encoding, presenting another limitation. This raises the question: \textit{Is it possible to design a scalable and efficient node-wise filter?}

In this work, we provide a positive answer to this question. We introduce polynomial attention \textbf{PolyAttn}, an attention-based node-wise filter. Our approach begins with the formulation of \textit{polynomial node tokens}, a concept analogous to tokens in a sentence or patches in an image. Through recursive calculation, these tokens are computed efficiently with complexity $O(K|E|)$, where $K$ signifies the truncated order of the polynomial basis which is often small in practice, and $|E|$ represents the number of edges in the graph. Once computed, these polynomial tokens can be reused across all training and inference stages, thereby enhancing efficiency. By applying attention to the $K+1$ polynomial tokens associated with each node, PolyAttn is able to learn node-wise filters in a mini-batch manner efficiently, resulting in a total complexity of $O((K+1)^2N)$, which ensures scalability. Additionally, the utilization of shared attention networks addresses issues related to excessive parameters and lack of generalization, while still maintaining the flexibility of node-wise filters. Through both theoretical and empirical analyses, we demonstrate that PolyAttn provides superior expressiveness over node-unified filters. Building upon this foundation of node-wise filters, we further develop the whole model \textbf{PolyFormer}.
% , which performs as a scalable and expressive Graph Transformer.

\textbf{In the lens of Graph Transformers}, PolyFormer is quite different from previous methods. PolyFormer implements attention on the polynomial tokens of each node, while former Graph Transformers generally calculate attention scores on nodes and then use the derived node representations for downstream tasks like node classification or graph regression \citep{architectureperspective_min2022transformer,attnbasedgnn_sun2023attention}. However, it is worth considering whether it is necessary to implement attention on node for both node-level and graph-level tasks. In natural language processing and computer vision, Transformer-based models mainly consider the interactions among tokens within a sentence or patches within an image, respectively, rather than implementing attention mechanisms between sentences or images \citep{transformer, bert, vit}. In other words, these attention mechanisms are typically employed on the sub-units that constitute the target object, rather than the target object itself. By capturing the information exchanges among these sub-units, the attention mechanism derives the representation of the target object. When it comes to Graph Transformers for node-level tasks, it is intuitive to develop an attention mechanism on tokens of each node rather than on nodes. 
In fact, for Graph Transformers, applying an attention mechanism on nodes in node-level tasks can potentially lead to several limitations. Firstly, calculating attention score on nodes leads to quadratic computational complexity with respect to the number of nodes \citep{transformer}, which poses challenges in terms of efficiency and scalability. Secondly, attending to all nodes in the graph neglect the sparse connectivity of the graph, which is a crucial inductive bias for graph structures. This has a detrimental impact on the performance of the Graph Transformer \citep{GraphTransformer_dwivedi2020generalization}. Although various works \citep{coarformer, ans-gt, nodeformer, goat_kong2023goat} have attempted to address these limitations through strategies such as sampling, coarsening, or adopting a more efficient attention mechanism, striking a balance between scalability and performance (e.g., accuracy) remains a challenge.

By implementing attention-based node-wise filter PolyAttn on the proposed polynomial token, \textbf{PolyFormer offers several advantages over existing graph transformers.}
\emph{Firstly, it processes attention within nodes rather than between nodes, enhancing scalability while keeping superior performance.} By adopting node tokens and implementing attention mechanisms on \(K+1\) tokens for each node instead of focusing on all nodes, the proposed model reduces computational complexity from \(O(N^2)\) to \(O((K+1)^2N)\), thereby improving scalability. What's more, PolyFormer supports mini-batch training, enhancing scalability further. 
% It's worth noting that although recent work NAGphormer \citep{nagphormer} has attempted to use information from various hops as units to represent each node and calculate attention on these units, the units designed based on the spatial domain, neglecting the spectral information, which compromises performance \citep{san}.
It's worth noting that recent work NAGphormer \citep{nagphormer} has attempted to represent each node using information from various hops as units to calculate attention on these units. However, these units are designed based on the spatial domain, neglecting the spectral information, which compromises performance \citep{san}.
In contrast, with the expressive node-wise filters PolyAttn, PolyFormer is able to maintain superior performance on large-scale graphs.   
\emph{Secondly, PolyFormer incorporates spectral information efficiently and demonstrates exceptional expressive power.} Previous Graph Transformers have highlighted the importance of spectral information for improving models' performance \citep{san, gps_rampavsek2022recipe, specformer}, but they rely on eigendecomposition of complexity $O(N^3)$, which is computationally demanding and memory-intensive. Our model, however, utilizes spectral information through polynomial approximation, offering both high expressiveness and efficiency. Extensive experiments have empirically validated PolyFormer's performance, efficiency, and scalability advantages.

We summarize the contributions of this paper as follows:
\vspace{-0.5mm}
\begin{itemize}
% \item We introduce a node-wise filter via a tailored attention mechanism, named PolyAttn. With polynomial-based node tokens from the spectral domain, PolyAttn is both scalable and expressive. Utilizing the node token and PolyAttn, we propose PolyFormer, a scalable and expressive Graph Transformer designed for node-level tasks.
\item We introduce a node-wise filter through a tailored attention mechanism, termed PolyAttn. By leveraging polynomial-based node tokens derived from the spectral domain, PolyAttn achieves both scalability and expressiveness. Utilizing these node tokens in conjunction with PolyAttn, we propose PolyFormer, serving as a scalable and expressive Graph Transformer for node-level tasks.
\item Theoretically, we demonstrate that PolyAttn functions as a node-wise filter with the designed node token. We also illustrate that multi-head PolyAttn serves as a multi-channel filter. Moreover, we explore the computational complexity tied to the proposed node token, PolyAttn and PolyFormer.
\item Comprehensive experiments validate that PolyAttn possesses greater expressive power than node-unified filters. Building on PolyAttn, PolyFormer achieves a desirable balance between expressive power and scalability. It demonstrates superior performance on both homophilic and heterophilic datasets and is capable of handling graphs with up to 100 million nodes.
\end{itemize}

\section{Background}
% In this section, we introduce the fundamental notations and concepts pertinent to graph signal filtering, Polynomial GNNs, and Transformers.

\subsection{Notations}
Consider an undirected graph $ G = (V, E) $, where $ V $ is the set of nodes and $ E $ is the set of edges. The adjacency matrix is denoted as $ \mathbf{A} \in \{0, 1\}^{N \times N} $, where $ \mathbf{A}_{ij} = 1 $ signifies the existence of an edge between nodes $ v_i $ and $ v_j $, and $ N $ is the total number of nodes in $G$. {The degree matrix $ \mathbf{D} $ is a diagonal matrix where $\mathbf{D}_{ii}=\sum\nolimits_{j}\mathbf{A}_{ij}$.}

% with each diagonal element $ \mathbf{D}_{ii} $ calculated as the sum of $ \mathbf{A}_{ij} $ for the corresponding node $ v_i $.
% The Graph Laplacian is defined as $ \mathbf{L} = \mathbf{D} - \mathbf{A} $, while its normalized version is $ \hat{\mathbf{L}} = \mathbf{I} - \hat{\mathbf{A}} = \mathbf{I} - \mathbf{D}^{-1/2}\mathbf{A}\mathbf{D}^{-1/2} $.

The normalized Laplacian of the graph is then defined as $\hat{\mathbf{L}} = \mathbf{I} - \hat{\mathbf{A}} = \mathbf{I} - \mathbf{D}^{-1/2}\mathbf{A}\mathbf{D}^{-1/2} $. In these equations, $ \mathbf{I} $ represents the identity matrix, $\hat{\mathbf{A}}$ denotes the normalized adjacency matrix, {which is obtained by scaling the adjacency matrix with the degree matrix.} It is well-established that $ \hat{\mathbf{L}} $ is a symmetric positive semidefinite matrix, allowing for decomposition as $ \hat{\mathbf{L}} = \mathbf{U}\mathbf{\Lambda}\mathbf{U}^{\top} =  \mathbf{U} \text{diag}(\lambda_0, \ldots, \lambda_{N-1})  \mathbf{U}^{\top}$. {Here, $\mathbf{\Lambda}$ is a diagonal matrix composed of real eigenvalues $\lambda_i \in [0,2]$, where $i \in \{0, \ldots, N-1\}$, and $\mathbf{U}$ consists of $N$ corresponding eigenvectors that are orthonormal, i.e., $\mathbf{U} = \{\boldsymbol{u}_0, \ldots, \boldsymbol{u}_{N-1}\}$.}

Further, we use $\boldsymbol{x} \in \mathbb{R}^{N}$ to denote the graph signal vector and use $\mathbf{X} \in \mathbb{R}^{N \times d}$ to denote the graph signal matrix or, equivalently, node feature matrix. The term $\boldsymbol{y} \in \mathbb{R}^N$ is used to denote the node label of the graph. Typically, in a \textit{homophilic} graph, node labels between neighbors tend to be the same, whereas in a \textit{heterophilic} graph, labels between neighbors tend to be different.

\subsection{Graph Filter}
Graph filter serves as a crucial concept in the field of graph signal processing \citep{GSP_isufi2022graph}.
% In graph filtering, graph signals are transformed from the spatial domain into the spectral domain and filters signals in the spectral domain, which is analogous to the Fourier transformation transforms signals from the time domain into the frequency domain and extracts signals in the frequency domain.
In graph filtering, graph signals are transformed from the spatial domain to the spectral domain, analogous to how the Fourier transformation converts signals from the time domain to the frequency domain. This process allows for filtering signals on the spectral domain, facilitating the extraction of specific signal components.

\textbf{Graph Signal Filtering.} 
Formally, given an original graph signal \( \boldsymbol{x} \in \mathbb{R}^{N} \), the filtered signal \( \boldsymbol{z} \in \mathbb{R}^{N} \) is obtained through a graph filtering operation in the spectral domain, expressed as:
\begin{equation}\label{graph_filter_1}
\boldsymbol{z} = \mathbf{U}h(\mathbf{\Lambda})\mathbf{U}^{\top}\boldsymbol{x}.
\end{equation}

In the equation above, graph signals \( \boldsymbol{x} \) are first projected into the spectral domain via the \textit{Graph Fourier Transform}: \( \hat{\boldsymbol{x}} = \mathbf{U}^{\top}\boldsymbol{x} \in \mathbb{R}^{N} \), employing the basis of frequency components \( \mathbf{U} = \{\boldsymbol{u}_0, \ldots, \boldsymbol{u}_{N-1}\} \). Here, \( \hat{\boldsymbol{x}} \) denotes the \textit{frequency response} of the original signal on the basis of frequency components in the spectral domain. The graph filter \( h(\cdot) \) then modulates the intensity of each frequency component through \( h(\mathbf{\Lambda})\mathbf{U}^{\top}\boldsymbol{x} \). Subsequently, the filtered signal is transformed back into the spatial domain using the \textit{Inverse Fourier Transform}, that is, \( \boldsymbol{z} = \mathbf{U}h(\mathbf{\Lambda})\mathbf{U}^{\top}\boldsymbol{x} \). These processes are also applicable to the node feature matrix \( \mathbf{X} \in \mathbb{R}^{N \times d} \):
\begin{equation}\label{graph_filter_2}
\mathbf{Z}=\mathbf{U}h(\mathbf{\Lambda})\mathbf{U}^{\top}\mathbf{X}.
\end{equation}
% It is worth noting that learning graph filters on $\mathbf{\Lambda}$ necessitates eigendecomposition, the process decomposite the Laplacian matrix in the form of $\mathbf{L}= \mathbf{U}\mathbf{\Lambda}\mathbf{U}^{\top}$, which has a time complexity of $ O(N^3)$, hinder the efficiency and scalability on large-scale graph.
It is worth noting that learning graph filters on $\mathbf{\Lambda}$ necessitates \textit{Laplacian eigendecomposition}, a process that decomposes the Laplacian matrix in the form of $\mathbf{L} = \mathbf{U}\mathbf{\Lambda}\mathbf{U}^{\top}$, which has a time complexity of $O(N^3)$. This significantly hinders efficiency and scalability on large-scale graphs.

\textbf{Node-wise and Channel-wise Filters.} 
Considering node signal matrix \( \mathbf{X} \in \mathbb{R}^{N \times d} \), graph filters can be expanded into more flexible and expressive forms: \textit{channel-wise} and \textit{node-wise}.
Specifically, \( h(\cdot) \) is considered {channel-wise} if there exists a corresponding \( h_{(j)}(\cdot) \) for each signal channel \( \mathbf{X}_{:,j} \), where \( j \in \{0, \ldots, d-1\} \). AdaGNN \citep{adagnn} learns different filters for each feature channel. Similarly, JacobiConv \citep{jacobi} and OptBasisGNN \cite{optbasis_guo2023graph} also implement multi-channel filters on their proposed bases. Conversely, graph filters are {node-wise} when \( h(\cdot) \) is tailored for individual nodes, denoted as \( h^{(i)}(\cdot) \) for node \( v_i \). DSF \citep{dsf_guo2023graph} implements a learnable network on the positional encoding to derive node-wise polynomial coefficients and shows enhanced performance of node-wise filters. However, the high overhead of initializing and updating the positional encoding poses a challenge to extending this approach to large-scale graphs.
% However, the significant overhead associated with initializing and updating the position matrix poses a challenge to extending this approach to large-scale graphs.

\textbf{Polynomial GNNs.} 
To alleviate the computational burden associated with eigendecomposition, recent studies have introduced polynomial GNNs to approximate \( h(\mathbf{\Lambda}) \) based on various polynomial bases. ChebNet \citep{Chebnet} adopts the Chebyshev basis to approximate the filtering operation, while GCN \citep{gcn_kipf2016semi} simplifies the ChebNet by limiting the Chebyshev basis to the first order. GPRGNN \citep{chien2021GPR-GNN} learns filters based on the Monomial basis, while BernNet \cite{bernnet} utilizes the Bernstein basis, providing enhanced interpretability. Furthermore, ChebNetII \citep{chebnetii_he2022convolutional} revisits the Chebyshev basis and constrains the coefficients through Chebyshev interpolation, showing minimax polynomial approximation property of truncated Chebyshev expansions. JacobiConv \cite{jacobi} further filters based on the family of Jacobi polynomial bases, with the specific basis determined by hyperparameters. \cite{optbasis_guo2023graph} introduces FavardGNN, which employs a learnable orthonormal basis for a given graph and signal, and OptBasisGNN, which utilizes an optimal basis with superior convergence properties.

Generally, using a specific polynomial basis, the approximated filtering operation can be represented as:
\begin{equation}
\mathbf{Z} = \mathbf{U}h(\mathbf{\Lambda})\mathbf{U}^{\top} \mathbf{X} \approx \sum_{k=0}^{K} \alpha_k g_k(\mathbf{P})\mathbf{X},
\end{equation}
where $ \alpha_k $ are the polynomial coefficients for all nodes, $ g_k(\cdot), k \in \{0,\ldots, K\} $ denotes a series polynomial basis of truncated order $ K $, and $\mathbf{P}$ refers to either the normalized adjacency matrix $ \hat{\mathbf{A}} $ or the normalized Laplacian matrix $ \hat{\mathbf{L}} $. For example, the filtering operation of GPRGNN~\citep{chien2021GPR-GNN} is $\mathbf{Z} = \sum\nolimits_{k=0}^{K} \alpha_k \hat{\mathbf{A}}^{k}\mathbf{X}$, which uses the Monomial basis.

% ----------------------------------
\subsection{Transformer}
The Transformer architecture \citep{transformer} is a powerful deep learning model that has shown a significant impact in multiple fields, including natural language processing \citep{bert, gpt3}, computer vision \citep{vit, swintransformer_liu2021swin, GTinCV_FCS}, audio applications \citep{speechtransformer_dong2018speech, conformer} and even graph learning \citep{architectureperspective_min2022transformer,attnbasedgnn_sun2023attention}.

\textbf{Attention Mechanism.}
% Formally, a Transformer model consists of multiple encoders and decoders. These encoder (decoder) blocks consist of attention modules and feedforward modules. The critical component is the \textit{attention mechanism}, which calculates pair-wise interactions between input tokens, allowing to capture global patterns.
The critical component of the Transformer is its \textit{attention mechanism}, which calculates pair-wise interactions between input tokens.
% Typiccaly, for an input matrix $ \mathbf{X} = [\boldsymbol{x}_1, \ldots, \boldsymbol{x}_n]^{\top} \in \mathbb{R}^{n \times d}$ with $n$ tokens, the attention mechanism transforms $ \mathbf{X} $ into the Query $ \mathbf{Q} $, the Key matrix $ \mathbf{K} $, and the Value matrix $ \mathbf{V} $ using learnable projection matrices $ \mathbf{W}_Q \in \mathbb{R}^{d \times d'} $, $ \mathbf{W}_K \in \mathbb{R}^{d \times d'} $, and $ \mathbf{W}_V \in \mathbb{R}^{d \times d'} $ as:
Typically, for an input matrix $\mathbf{X} = [\boldsymbol{x}_1, \ldots, \boldsymbol{x}_n]^{\top} \in \mathbb{R}^{n \times d}$ with $n$ tokens, the attention mechanism transforms $\mathbf{X}$ into $\mathbf{Q}$, $\mathbf{K}$, $\mathbf{V}$ via learnable projection matrices $\mathbf{W}_Q \in \mathbb{R}^{d \times d'}$, $\mathbf{W}_K \in \mathbb{R}^{d \times d'}$, and $\mathbf{W}_V \in \mathbb{R}^{d \times d'}$ as:
\begin{equation}
\mathbf{Q} = \mathbf{X}\mathbf{W}_Q, \quad \mathbf{K} = \mathbf{X}\mathbf{W}_K, \quad \mathbf{V} = \mathbf{X}\mathbf{W}_V.
\end{equation}
The output of the attention mechanism is then computed as:
\begin{equation}
\mathbf{O} = \text{softmax}\left(\frac{\mathbf{Q}\mathbf{K}^{\top}}{\sqrt{d}}\right)\mathbf{V}.
\end{equation}
% This attention mechanism can be executed multiple times to produce a \textit{multi-head attention} mechanism. It is worth noting that the complexity of attention is quadratic with the input sequence length, i.e. $O(n^2)$.
This attention mechanism can be executed multiple times to implement a \textit{multi-head attention} mechanism. It is important to note that the complexity of the attention mechanism is quadratic with respect to the input sequence length, denoted as $O(n^2d')$.

\textbf{Graph Transformer.}
The Transformer architecture, adapted to graph learning and termed Graph Transformers, has garnered significant attention in recent years \citep{architectureperspective_min2022transformer,attnbasedgnn_sun2023attention}. 
% When applying Transformers to the graph domain, it typically treats nodes of the graph as input tokens for the model. However, computing overheard of attention which quadratic to the number of nodes restricting the \textbf{scalability of Graph Transformers}. 
% When adapting Transformers to the graph domain, nodes of the graph are typically treated as input tokens for the model. However, the computational overhead of the attention mechanism, which is quadratic with respect to the number of nodes, restricts the application of Graph Transformers on large-scale graphs.
When adapting Transformers to graph learning, nodes of the graph are typically treated as input tokens for the model. However, the computational overhead of the attention mechanism, which scales quadratically with the number of nodes, restricts the application of Graph Transformers on large-scale graphs.

Various solutions have been proposed to address the scalability limitation of Graph Transformers.
(1) Adopting Efficient attention mechanisms enhances scalability. Nodeformer \cite{nodeformer} employs existing efficient attention \cite{performer} with Gumbel-Softmax \cite{gumblesoftmax} for large-scale graphs, while SGFormer \cite{SGFormer} optimizes scalability by substituting softmax attention with proposed linear attention.
% (2) Node coarsening, as another strategy, involves simplifying graph structures. Coarformer \cite{coarformer} employs graph coarsening algorithms \cite{coarsen1_loukas2019graph,coarsen2_ron2011relaxation} to gain coarsened nodes, enabling attention within and across coarsened and local nodes. GOAT \cite{goat_kong2023goat} utilizes the exponential moving average strategy and K-Means \cite{kmeans_jain1988algorithms} to integrate global information into codebooks and implement attention on them, and \cite{ans-gt} also incorporates coarsened nodes for attention processes.
(2) Node coarsening, as another strategy, involves simplifying graph structures. Coarformer \cite{coarformer} employs graph coarsening algorithms \cite{coarsen1_loukas2019graph,coarsen2_ron2011relaxation} to obtain coarsened nodes, thereby enabling attention within and across coarsened and local nodes. GOAT \cite{goat_kong2023goat} utilizes the exponential moving average strategy and K-Means \cite{kmeans_jain1988algorithms} to integrate global information into codebooks and implement attention on them. Additionally, ANS-GT \cite{ans-gt} utilizes coarsened nodes for attention processes.
% (3) Sampling strategies, like those used by ANS-GT \cite{ans-gt} with several sampling heuristics (e.g., 1-hop neighbors, Personalized PageRank), reduce the computational demand by selecting specific nodes for attention calculations. Despite these methods, balancing scalability with performance in Graph Transformers is an ongoing challenge.
(3) Sampling strategies, such as those employed by ANS-GT \cite{ans-gt} with various sampling heuristics (e.g., 1-hop neighbors, Personalized PageRank), reduce the computational demand by selecting specific nodes for attention calculations. Despite these approaches, balancing scalability with performance in Graph Transformers remains an ongoing challenge.

% \newpage
% \clearpage
% \clearpage

\section{PolyFormer}

% In this section, we introduce our proposed node-wise filter PolyAttn and scalable graph transformer PolyFormer. First, we define the concept of node tokens based on polynomial bases.
In this section, we introduce our proposed node-wise filter PolyAttn and the whole model PolyFormer, which serves as a scalable and expressive Graph Transformer. First, we define the concept of node tokens based on polynomial bases.
% in the spectral domain. 
Utilizing these node tokens, we describe our attention-based node-wise filter and provide an overview of the whole model. Finally, we analyze the computational complexity of our methods and illustrate its relationship with graph filters.

\subsection{Polynomial Token}
Analogous to sentence tokenization in natural language processing, we introduce polynomial tokens.
% for nodes to enhance Graph Transformer scalability on node-level tasks.
% Analogous to the tokenization of sentences in natural language processing, we introduce the concept of polynomial tokens for nodes. This is aimed at enhancing the scalability of the Graph Transformer for node-level tasks.

\begin{definition}\label{de:polynomial_token}
{\rm \textbf{(Polynomial Token)}} For any node $v_i$ in a graph $G = (V, E)$, the polynomial token of the node is defined as $\boldsymbol{h}^{(i)}_{k} = \left( g_{k}(\mathbf{P})\mathbf{X}\right)_{i,:} \in \mathbb{R}^d, k \in \{0, \ldots, K\}$, where $g_k(\cdot)$ represents a polynomial basis of order $k$, $\mathbf{P}$ is either $\hat{\mathbf{A}}$ or $\hat{\mathbf{L}}$, and $\mathbf{X}$ represents the node features.
\end{definition}

% In this work, we employ Monomial and Chebyshev bases for polynomial tokens. These choices offer ease of implementation in comparison to more complex polynomial bases like Bernstein or Jacobi. Besides, The Monomial basis provides a clear spatial interpretation, with $\boldsymbol{h}^{(i)}_{k} = ( \hat{\mathbf{A}
% }^k \mathbf{X} )_{i,:}$ representing the $k$-hop neighborhood information for node $v_i$. Meanwhile, the Chebyshev base exhibits excellent fitting capabilities \citep{geddes1978near}. 
In this work, we employ Monomial, Bernstein, Chebyshev, and the optimal bases for polynomial tokens. These choices offer ease of implementation compared to some more complex polynomial bases such as Jacobi basis. Additionally, these bases provide good properties. the Monomial basis provides a clear spatial interpretation, with $\boldsymbol{h}^{(i)}_{k} = ( \hat{\mathbf{A}}^k \mathbf{X} )_{i,:}$ representing the information of the $k$-hop neighborhood from node $v_i$, while the Bernstein basis coefficients are highly correlated with the spectral property of the filer, providing good interpretability. The Chebyshev basis exhibits excellent fitting capabilities \citep{geddes1978near}, while the optimal basis provides the best converge property \cite{optbasis_guo2023graph}.
% Both bases can be computed recursively, as illustrated in Table \ref{recursive_computing_of_polynomial_tokens}.
Table \ref{recursive_computing_of_polynomial_tokens} illustrates the computing process of polynomial tokens for all nodes in the graph, where $\mathbf{H}_k = [\boldsymbol{h}_k^{(0)}, \ldots, \boldsymbol{h}_k^{(N-1)}]^{\top} \in \mathbb{R}^{N \times d}$ denotes the matrix consisting of polynomial tokens of order $k$ for $k \in \{0, \ldots, K\}$.

\begin{figure*}[t] 
\centering
% \resizebox{0.74\textwidth}{!}{%
\resizebox{0.78\textwidth}{!}{%
\includegraphics[width=1.0\textwidth]{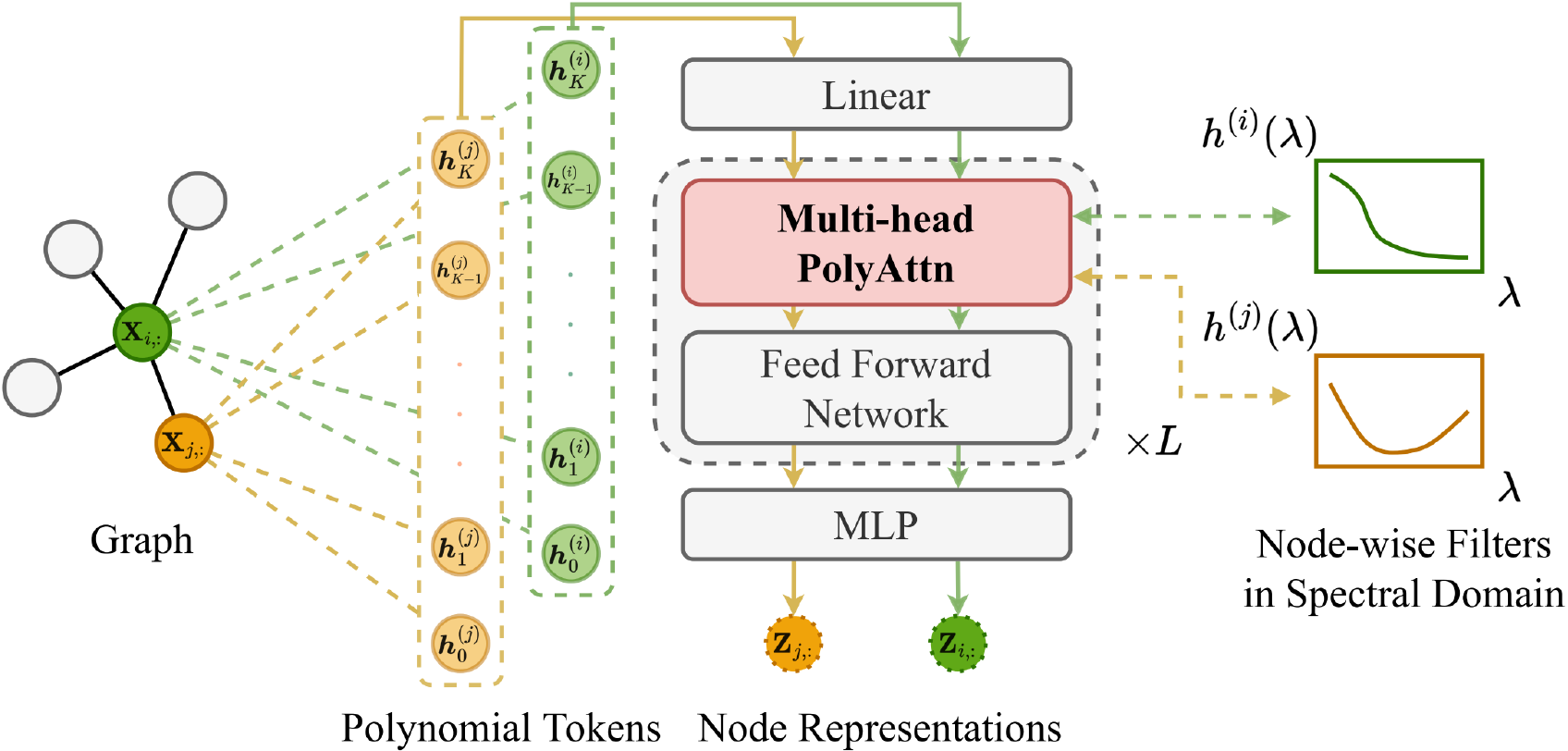}
}
\caption{\textbf{Illustration of the proposed PolyFormer.} For a given graph, polynomial tokens for each node are computed. These tokens are subsequently processed by PolyFormer, which consists of \( L \) blocks. Notably, with the defined polynomial token, PolyAttn within each block functions as a node-wise filter in the spectral domain, adaptively learning graph filter specific to each node.}
\label{model_illustration}
% \vspace{-5mm}
\end{figure*}

% KDD table v1
\begin{table}[t]
  \centering
  % \vspace{-3mm}
  \caption{Computing process of polynomial tokens for different bases.}
  % \vspace{-2mm}
    % \resizebox{0.5\textwidth}{!}{%
  \resizebox{\linewidth}{!}{
    \begin{tabular}{ccc}
    \toprule
    \textbf{Type} & \textbf{Basis $g_k$} & \textbf{Token}  \\
    \midrule
    Monomial & $g_k(\hat{\mathbf{A}}) = \hat{\mathbf{A}} g_{k-1}(\hat{\mathbf{A}}), g_0(\hat{\mathbf{A}}) = \mathbf{I}$ %\tablefootnote{$g_0(\hat{\mathbf{A}}) = \mathbf{I}$} 
    & $\mathbf{H}_k = g_k(\hat{\mathbf{A}}) \mathbf{X}$ \\

    Bernstein & $g_k(\hat{\mathbf{L}}) = \frac{1}{2^K} \binom{K}{k}(2\mathbf{I}- \hat{\mathbf{L}})^{K-k}(\hat{\mathbf{L}})^k$ & $\mathbf{H}_k = g_k(\hat{\mathbf{L}}) \mathbf{X}$ \\
    
    Chebyshev & $g_k(\hat{\mathbf{L}}) = 2 \hat{\mathbf{L}} g_{k-1}(\hat{\mathbf{L}}) - g_{k-2}(\hat{\mathbf{L}})$ \tablefootnote{\(g_0(\hat{\mathbf{L}})=\mathbf{I},g_1(\hat{\mathbf{L}})=\hat{\mathbf{L}}\)} & $\mathbf{H}_k = g_k(\hat{\mathbf{L}}) \mathbf{X}$ \\

    % Optimal & $g_k(\hat{\mathbf{A}}, \mathbf{X}) = \delta_{k-1} g_{k-1}(\hat{\mathbf{A}}, \mathbf{X}) + \delta_{k-2} g_{k-2}(\hat{\mathbf{A}}, \mathbf{X})$\tablefootnote{$g_{0}(\hat{\mathbf{A}}, \mathbf{X})=\mathbf{0},g_{1}(\hat{\mathbf{A}}, \mathbf{X})=\mathbf{X}, \delta_i$ is determined by both $\hat{\mathbf{A}}, \mathbf{X}$}  & $\mathbf{H}_k = g_k(\hat{\mathbf{A}}, \mathbf{X})$ \\

    % Optimal & $\beta_{k} g_k(\hat{\mathbf{A}}) = (\hat{\mathbf{A}} - \gamma_{k-1} \mathbf{I}) g_{k-1}(\hat{\mathbf{A}}) - \beta_{k-1} g_{k-2}(\hat{\mathbf{A}})$ \tablefootnote{$g_{-1}(\hat{\mathbf{A}})=\mathbf{0},g_{0}(\hat{\mathbf{A}})=\mathbf{I}/||\mathbf{X}||$}  & $\mathbf{H}_k = g_k(\hat{\mathbf{A}}) \mathbf{X}$ \\

    Optimal & $\beta_{k} g_k(\hat{\mathbf{A}}) = (\hat{\mathbf{A}} - \gamma_{k-1} \mathbf{I}) g_{k-1}(\hat{\mathbf{A}}) - \beta_{k-1} g_{k-2}(\hat{\mathbf{A}})$ \tablefootnote{$g_{-1}(\hat{\mathbf{A}})=\mathbf{0},g_{0}(\hat{\mathbf{A}})=\mathbf{I}/||\mathbf{X}||$. According to \cite{optbasis_guo2023graph}, the optimal bases differs on each channel of $\mathbf{X}$. $\gamma_k$ and $\beta_k$ here, and thus the $k$-order output basis, are determined by $\mathbf{A}$ and $\mathbf{X}_{:,j}$ specifically.}  & $(\mathbf{H}_k )_{:,j} = (g_k(\hat{\mathbf{A}})) \mathbf{X}_{:,j}$ \\

    \bottomrule
    \end{tabular}%
    }
  \label{recursive_computing_of_polynomial_tokens}%
\end{table}%

The adoption of polynomial tokens presents several distinct advantages. Firstly, these tokens can be computed efficiently. As shown in Table \ref{recursive_computing_of_polynomial_tokens}, most proposed polynomial tokens can be computed recursively, which is an efficient manner. More detailed complexity analysis is illustrated in \ref{complexity}. 
% Meanwhile, although the Bernstein token computation is quadratic in relation to \(K\), it remains efficient. 
Once computed, they can be reused across each epoch during both the training and inference phases, leading to substantial reductions in computational time and memory usage. 
Furthermore, by incorporating the normalized adjacency or Laplacian matrix $\mathbf{P}$ into the computational process, graph topology information is integrated into node tokens. This integration eliminates the necessity for additional positional or structural encodings, such as Laplacian eigenvectors, thereby enhancing the model's efficiency. Finally, the node-wise independence of these polynomial tokens allows for mini-batch training, enabling our model to scale to graphs with up to 100 million nodes.
% Finally, the inherent node-wise independence of these polynomial tokens enables mini-batch training, allowing us to scale the model to a graph with 100 million nodes.

\subsection{PolyAttn and PolyFormer}
Given the polynomial tokens associated with each node, PolyFormer employs an expressive attention-based node-wise filter PolyAttn to generate node representations. Firstly, we introduce the proposed attention mechanism PolyAttn, which is tailored for polynomial tokens and performs as a node-wise filter. Subsequently, we detail the comprehensive architecture of PolyFormer.

\textbf{PolyAttn.}
In this section, we first detail the process of the proposed PolyAttn for a given node \(v_i\). We define the token matrix \(\mathbf{H}^{(i)}\) for node \(v_i\) as \(\mathbf{H}^{(i)} = [\boldsymbol{h}^{(i)}_0,\ldots, \boldsymbol{h}^{(i)}_K]^{\top} \in \mathbb{R}^{(K+1) \times d}\). Initially, the value matrix $\mathbf{V}$ is initialized with the token matrix $\mathbf{H}^{(i)}$. 
Subsequently, an order-specific multi-layer perceptron (MLP\(_j\)) maps the \(j\)-th order token \(\boldsymbol{h}^{(i)}_j = \mathbf{H}_{j,:}^{(i)}\) into a hidden space. This mapping captures unique contextual information for each order of polynomial tokens.

Then the query matrix \(\mathbf{Q}\) and the key matrix \(\mathbf{K}\) are obtained by projecting \(\mathbf{H}^{(i)}\) through the learnable matrices \(\mathbf{W}_Q\) and \(\mathbf{W}_K\), respectively. These matrices compute the attention scores. Notably, our attention mechanism utilizes the hyperbolic tangent function \(\tanh(\cdot)\) rather than the softmax function commonly used in the vanilla Transformer~\citep{transformer}. 
% This is because the softmax function limits the expressive capability of PolyAttn when it functions as a node-wise graph filter. Further clarification is provided in Proposition \ref{proposition0}. Additionally, a node-shared attention bias $\boldsymbol{\beta} \in \mathbb{R}^{K+1}$ is introduced to strike a balance between node-specific and global patterns. Finally, the computed attention scores $\mathbf{S}$ are used to multiply with the value matrix $\mathbf{V}$, resulting in the final output representations. The pseudocode of PolyAttn is provided as Algorithm~\ref{finalpsecode}. 
This modification addresses the limitations of the softmax function in expressing the capabilities of PolyAttn as a node-wise graph filter, as detailed in Proposition \ref{proposition0}. Additionally, we introduce a node-shared attention bias \(\boldsymbol{\beta} \in \mathbb{R}^{K+1}\) to balance node-specific and global patterns. The attention scores \(\mathbf{S}\) are then used to update the value matrix \(\mathbf{V}\), yielding the final output representations. The pseudocode for PolyAttn is outlined in Algorithm~\ref{finalpsecode}.
For enhanced expressive power, multi-head PolyAttn is utilized in practice, with further details available in Appendix~\ref{implementation_details}.

\textbf{PolyFormer.} 
Building upon the attention mechanism designed for polynomial tokens, we introduce the whole model PolyFormer. As illustrated in Figure \ref{model_illustration}, PolyFormer block is described by the following equations:

\begin{align}
    \mathbf{H'}^{(i)} &= \text{PolyAttn}\left(\text{LN}\left(\mathbf{H}^{(i)}\right)\right) + \mathbf{H}^{(i)}, \\
    \mathbf{H}^{(i)}  &= \text{FFN}\left(\text{LN}\left(\mathbf{H'}^{(i)}\right)\right) + \mathbf{H'}^{(i)}.
\end{align}

Here, LN denotes Layer Normalization, which is implemented before PolyAttn \citep{layernorm_xiong2020layer}. FFN refers to the Feed-Forward Network. Upon obtaining the token matrix $\mathbf{H}^{(i)} \in \mathbb{R}^{(K+1) \times d}$ for node $v_i$ through $L$ PolyFormer blocks, the final representation $\mathbf{Z}_{i,:} \in \mathbb{R}^{c}$ of node $v_i$ is computed as:

\begin{equation}
\mathbf{Z}_{i,:} = \sigma\left( \left( \sum_{k=0}^{K}  \mathbf{H}^{(i)}_{k,:} \right) \mathbf{W}_{1}\right)\mathbf{W}_{2},
\end{equation}

where $\sigma$ denotes the activation function. The matrices $\mathbf{W}_1 \in \mathbb{R}^{d \times d'}$ and $\mathbf{W}_2 \in \mathbb{R}^{d' \times c}$ are learnable, with $d, d'$ representing the hidden dimensions and $c$ representing the number of node classes.

% -----------------------------------------------------------------
\begin{algorithm}[ht]
\DontPrintSemicolon
\SetKwInput{Input}{Input}
\SetKwInput{Output}{Output}

\Input{Token matrix for node $v_i$: $\mathbf{H}^{(i)}=[\boldsymbol{h}^{(i)}_0,\ldots, \boldsymbol{h}^{(i)}_K]^{\top} \in \mathbb{R}^{(K+1) \times d}$}

\Output{New token matrix for node $v_i$: $\mathbf{H'}^{(i)} \in \mathbb{R}^{(K+1)\times d}$}

\Parameter{
% $\beta$, $\gamma$, $\alpha$
% \textbf{Learnable parameters:}
Projection matrix $\mathbf{W}_Q$,$\mathbf{W}_K\in\mathbb{R}^{d \times d'}$,\\
order-wise MLP$_j(j =0, \ldots, K)$, \\ 
attention bias $\boldsymbol{\beta}\in \mathbb{R}^{K+1}$ \;
}
Initialize $\mathbf{V}$ with $\mathbf{H}^{(i)}$\;

\For{j = $0$ \textnormal{\textbf{to}} $K$}{
   $\mathbf{H}^{(i)}_{j,:} \leftarrow \textnormal{MLP}_j(\mathbf{H}^{(i)}_{j,:})$\;
}

$\mathbf{Q} \leftarrow \mathbf{H}^{(i)}  \mathbf{W}_Q$ via projection matrix $\mathbf{W}_Q$; $\mathbf{K} \leftarrow \mathbf{H}^{(i)}  \mathbf{W}_K$ via projection matrix $\mathbf{W}_K$\;

Compute attention scores $\mathbf{S} \leftarrow  \text{tanh}(\mathbf{Q}\mathbf{K}^{\top}) \odot \mathbf{B}$, where  $\mathbf{B}_{ij} = {\boldsymbol{\beta}}_j$\;

$\mathbf{H'}^{(i)} \leftarrow  \mathbf{S} \mathbf{V}$\;

\Return $\mathbf{H'}^{(i)}$ \quad  \# The  representation of node $v_i$ after PolyAttn is $\boldsymbol{Z}_{i,:} = \sum_{k=0}^{K} \mathbf{H'}^{(i)}_{k,:} \in \mathbb{R}^d$.

\caption{Pseudocode for PolyAttn} \label{finalpsecode}
% \vspace{-1mm}
\end{algorithm}
% \vspace{-2mm}
% -----------------------------------------------------------------

\subsection{Theoretical Analysis}
\subsubsection{Complexity} \label{complexity}
% \textbf{Preprocessing for Polynomial Tokens.}\label{complexity_1}
% \
% \newline
% \noindent
Here, we analyze the complexity of computing polynomial tokens, PolyAttn and PolyFormer.

\textbf{Computing for Polynomial Tokens.}
As previously discussed, the polynomial tokens of Monimial, Chenyshev and the optimal basis can be calculated recursively. Each iteration for all nodes involves sparse multiplication with a computational complexity of $O(|E|)$. Thus, the overall complexity is $O(K|E|)$, where $K$ is the truncated order of the polynomial tokens, and $|E|$ is the number of edges in the graph. It is worth noting that though Bernstein polynomial tokens are computed in the complexity of $O(K^2|E|)$, it is still efficient as $K$ is small in practice. Importantly, these polynomial tokens can be computed once and reused throughout the training and inference process.

\textbf{Complexity of PolyAttn and PolyFormer.}
% Let \( d \) denote the hidden dimension of polynomial tokens, and \( K \) represent the truncated order. For each node in the one layer PolyAttn, \( (K + 1) \) polynomial tokens are involved in attention computing, resulting in a complexity of \( O((K+1)^2d) \). With \( N \) nodes in the graph and \( L \) layers of attention mechanisms, the total time complexity is \( O(LN(K+1)^2d) \).
Let \( d \) denote the hidden dimension of polynomial tokens, and \( K \) represent the truncated order. In the context of one layer of PolyAttn, each node involves \( (K + 1) \) polynomial tokens in attention computation, resulting in a complexity of \( O((K+1)^2d) \). With \( N \) nodes in the graph and \( L \) layers of attention mechanisms, the total time complexity is \( O(LN(K+1)^2d) \).
% Let \( d \) denote the hidden dimension of the node token and \( K \) represent the truncated order of the polynomial tokens. For each node, there are \( (K + 1) \) polynomial tokens involved in the one layer PolyAttn, leading to a complexity of \( O((K+1)^2d) \) for a single node. With \( N \) nodes in the graph and $L$ layers attention mechanism, the total time complexity is \( O(LN(K+1)^2d) \). 
Notably, in practical situations where \( K \ll N \), this signifies a substantial reduction in computational complexity, especially when compared to the \( O(LN^2d) \) complexity of vanilla Transformer models.

\subsubsection{Connection to Spectral Filtering}

To understand the connection between PolyAttn and graph filters, we give the following theorem and propositions. All proofs are in Appendix~\ref{proof_all}. First, we formally propose that PolyAttn serves as a node-wise filter for polynomial tokens.

\begin{theorem} \label{theorem1}
% When taking polynomial tokens as input, PolyAttn operates as a node-wise filter. 
With polynomial tokens as input, PolyAttn operates as a node-wise filter. Specifically, for the representation $\mathbf{Z}_{i,:} = \sum_{k=0}^{K} \mathbf{H'}^{(i)}_{k,:}$ of node $v_i$ after applying PolyAttn: % calculated by  
\begin{equation}
\mathbf{Z}_{i,:} = \sum_{k=0}^{K} \mathbf{H'}^{(i)}_{k,:} = \sum_{k=0}^{K} \alpha_k^{(i)} \left(g_k\left(\mathbf{P}\right)\mathbf{X}\right)_{i,:}.
\end{equation}
Here, the coefficients $\alpha_k^{(i)}$ depend not only on the polynomial order $k$ but also on the specific node $v_i$. In other words, \textbf{PolyAttn performs a node-wise polynomial filter} on the graph signals.
\end{theorem}

\textbf{Theorem \ref{theorem1} builds the bridge between the proposed attention and node-wise filters.} Building on this, we further propose that the multi-head PolyAttn acts as a multi-channel filter.

\begin{proposition} \label{proposition1}
A multi-head PolyAttn with \( h \) heads can be interpreted as partitioning the node representation into \( {h} \) channel groups with dimension $d_h = \frac{d}{h}$ and applying filtering to each group separately. Formally:
\begin{equation}
    \mathbf{Z}_{i,p:q} = \sum_{k=0}^{K} \alpha^{(i)}_{(p,q)k} \left( g_k(\mathbf{P})\mathbf{X}\right)_{i,p:q} .
\end{equation}
% Here, \( \alpha^{(i)}_{(p,q)k} \) denotes the node-wise coefficient for channels \( p \) to \( q, \text{where} (p,q) = (j \times d_h, (j+1) \times d_h -1), j \in \{0, \ldots, h-1\} \).
% Here, \( \alpha^{(i)}_{(p,q)k} \) denotes the node-wise coefficient for channels \( p \) to \( q \), where \( (p,q) = (j \times d_h, (j+1) \times d_h -1), j \in \{0, \ldots, h-1\} \).
Here, \( \alpha^{(i)}_{(p,q)k} \) denotes the coefficient for order \( k \) on channels \( p \) to \( q \) of node \( v_i \)'s representation, where \( (p,q) = (j \times d_h, (j+1) \times d_h -1), j \in \{0, \ldots, h-1\} \).
% Here, \( \alpha^{(i)}_{(p,q)k} \) denotes the coefficient for order \( k \) on channels \( p \) to \( q \) of node \( v_i \)'s representation, where \( (p,q) = (j \times d_h, (j+1) \times d_h -1) \), with \( j \) ranging from \( 0 \) to \( h-1 \).
\end{proposition}

It is worth noting that our chosen activation function, $\tanh(\cdot)$, enables PolyAttn with more powerful expressiveness than the softmax function.

\begin{proposition} \label{proposition0}
For PolyAttn, which operates as a graph filter, the \textit{tanh} function endows it with enhanced expressiveness, whereas the softmax function can limit the expressive capability of PolyAttn.
\end{proposition}

\begin{table*}[t]
  \centering
  % \caption{Performance of PolyAttn on Synthetic Datasets ($R^2$ score / the sum of squared error).}
\caption{Performance of PolyAttn on synthetic datasets, presented as $R^2$ score (with values closer to 1 indicating better performance) and the sum of squared errors (with values closer to 0 indicating higher accuracy).}
  \vspace{-3mm}
  \resizebox{\textwidth}{!}{%
    \begin{tabular}{lcccccc}
    \toprule
    \textbf{Model (5k para.)} & \textbf{Mixed low-pass} & \textbf{Mixed high-pass} & \textbf{Mixed band-pass} & \textbf{Mixed rejection-pass} & \textbf{Low\&high-pass} & \textbf{Band\&rejection-pass} \\
    \midrule
    GCN   & 0.9953/2.0766 & 0.0186/39.6157 & 0.1060/14.0738 & 0.9772/10.9007 & 0.6315/86.8209 & 0.8823/128.2312 \\
    GAT   & 0.9954/2.0451 & 0.0441/38.5851 & 0.0132/14.0375 & 0.9775/10.7512 & 0.7373/61.8909 & 0.9229/83.9671 \\
    GPRGNN & 0.9978/0.9784 & 0.9806/0.7846 & 0.9088/1.2977 & 0.9962/1.8374 & 0.8499/35.3719 & 0.9876/13.4890 \\
    BernNet & 0.9976/1.0681 & 0.9808/0.7744 & 0.9231/1.0937 & 0.9968/1.5545 & 0.8493/35.5144 & 0.9875/13.6485 \\
    ChebNetII & 0.9980/0.8991 & 0.9811/0.7615 & {0.9492/0.7229} & {0.9982/0.8610} & 0.8494/35.4702 & 0.9870/14.1149 \\
    {PolyAttn (Mono)} & {0.9994/0.2550} & {0.9935/0.2631} & 0.9030/1.3798 & 0.9971/1.4025 & \underline{0.9997/0.0696} & \underline{0.9992/0.8763} \\
    {PolyAttn (Bern)} & \textbf{0.9998/0.0842} & 0.9972/0.1120 & 0.9809/0.2719 & 0.9993/0.3337 & 0.9852/3.4956 & 0.9882/12.8274 \\
    {PolyAttn (Opt)} & 0.9996/0.1922 & \textbf{0.9997/0.0103} & \textbf{0.9951/0.0701} & \underline{0.9995/0.2275} & 0.9978/0.5136 & 0.9992/0.8929 \\
    {PolyAttn (Cheb)} & \underline{0.9997/0.1467} & \underline{0.9960/0.0148} & \underline{0.9945/0.0782} & \textbf{0.9996/0.1949} & \textbf{0.9999/0.0118} & \textbf{0.9999/0.0416} \\
    \bottomrule
    \end{tabular}%
    }
  \label{result_SBM_PolyAtttn}%
  % \vspace{-1mm}
\end{table*}%

\begin{figure*}[th]
\centering
\begin{subfigure}{0.25\textwidth}
  \centering
  \includegraphics[width=\linewidth]{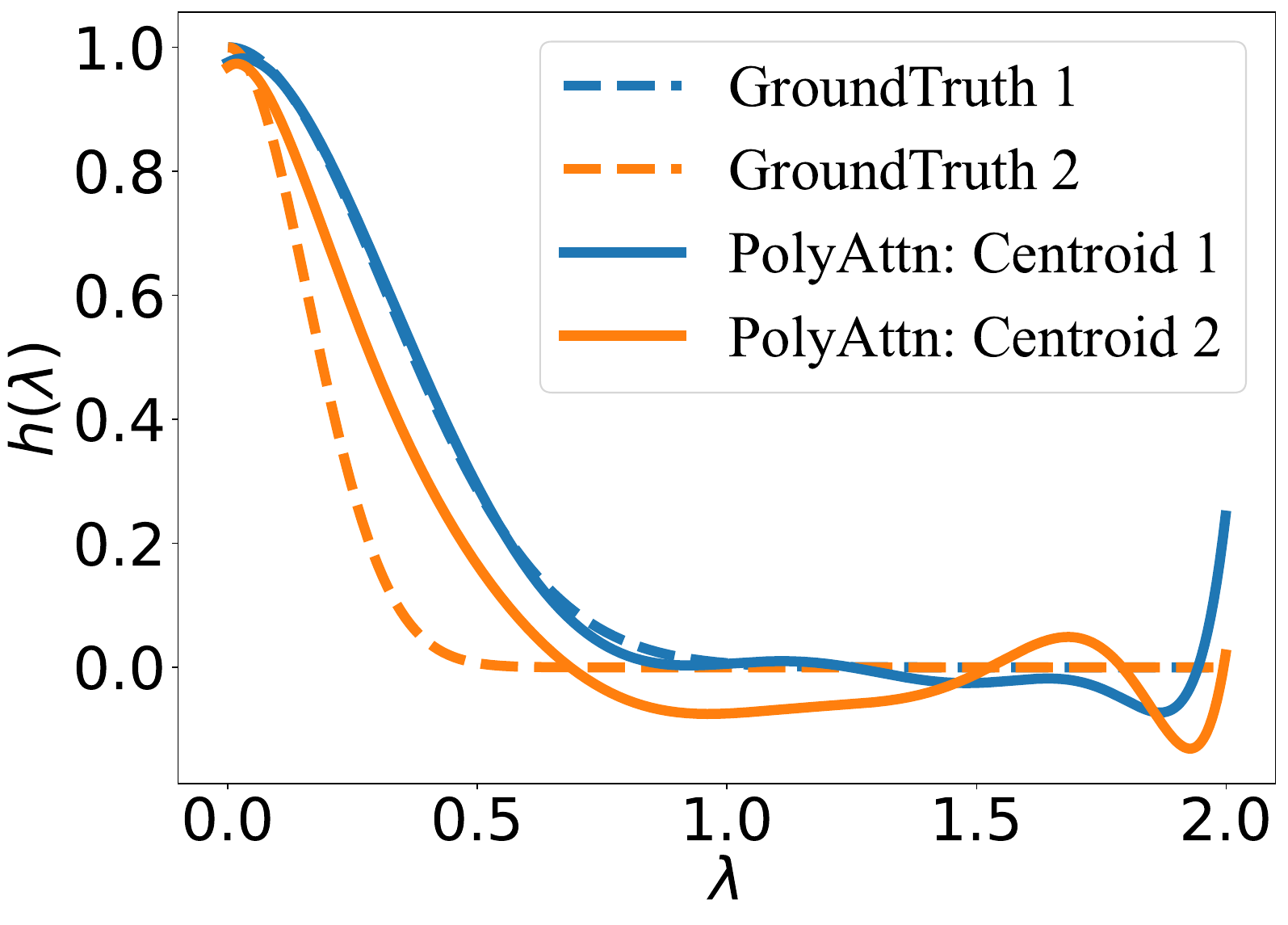}
  \caption{Mixed low-pass}
  \label{fig:image1}
\end{subfigure}%
\begin{subfigure}{0.25\textwidth}
  \centering
  \includegraphics[width=\linewidth]{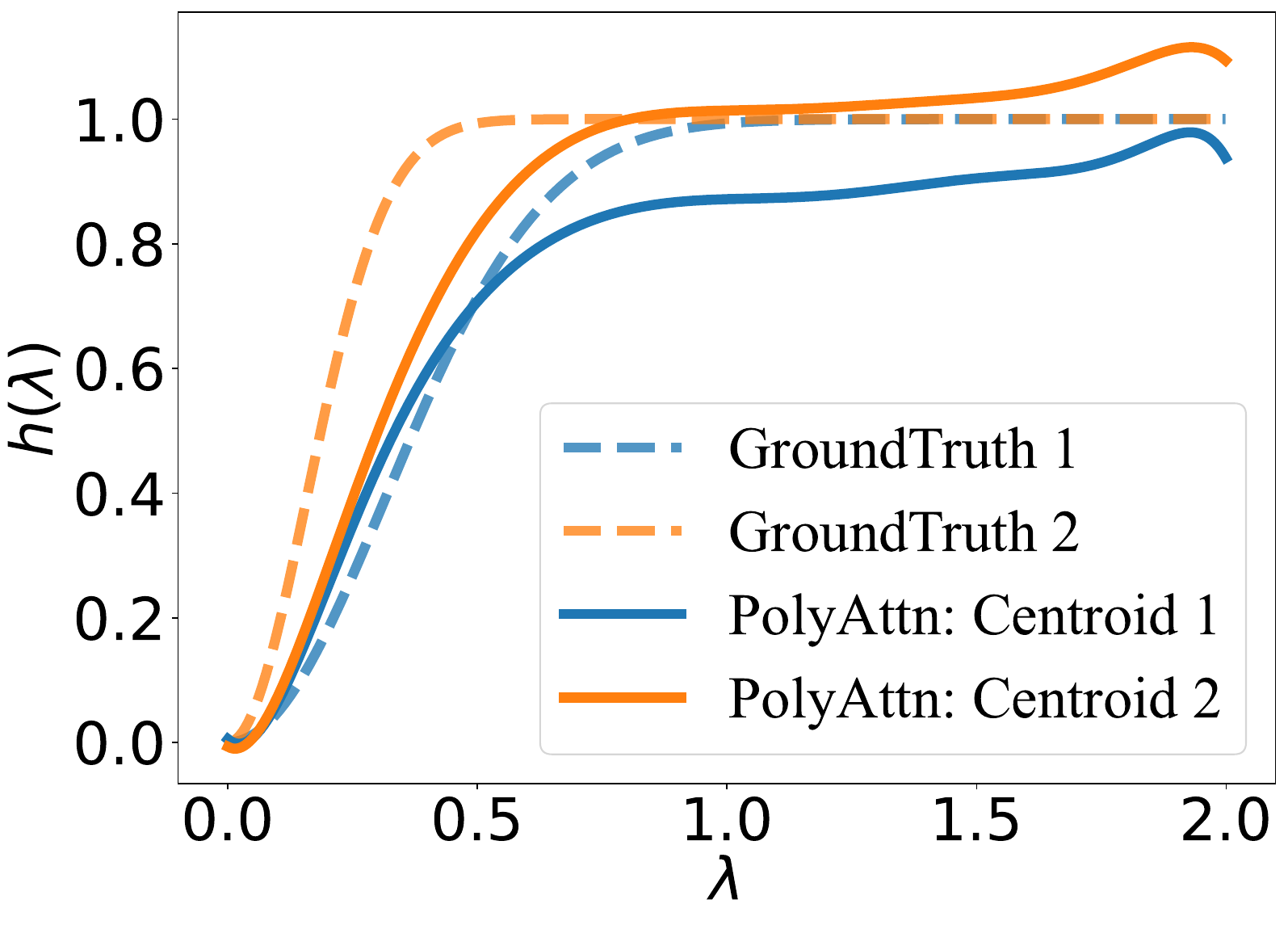}
  \caption{Mixed high-pass}
  \label{fig:image2}
\end{subfigure}%
\begin{subfigure}{0.258\textwidth}
  \centering
  \includegraphics[width=\linewidth]{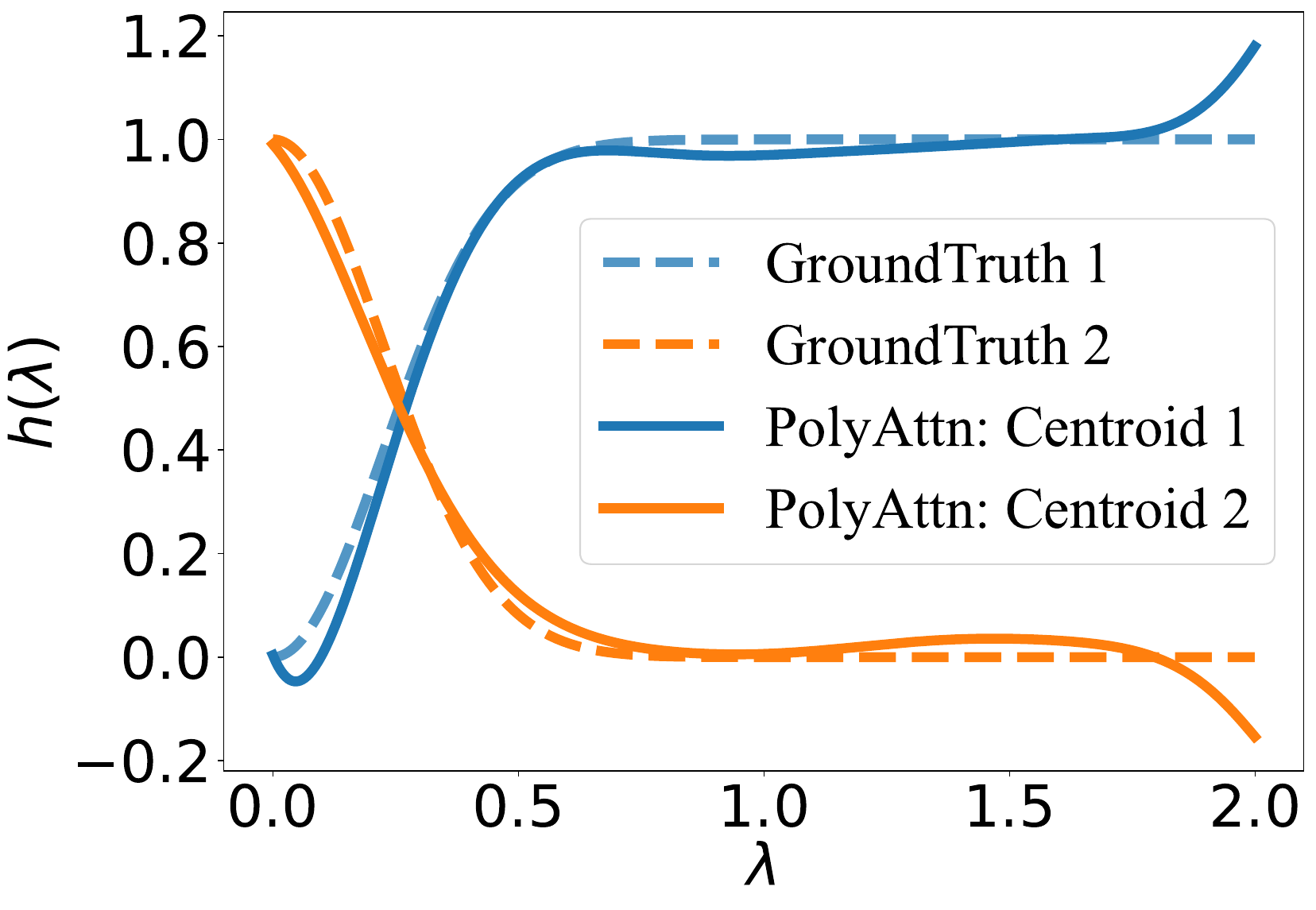}
  \caption{Low-pass \& high-pass}
  \label{fig:image3}
\end{subfigure}%
\begin{subfigure}{0.25\textwidth}
  \centering
  \includegraphics[width=\linewidth]{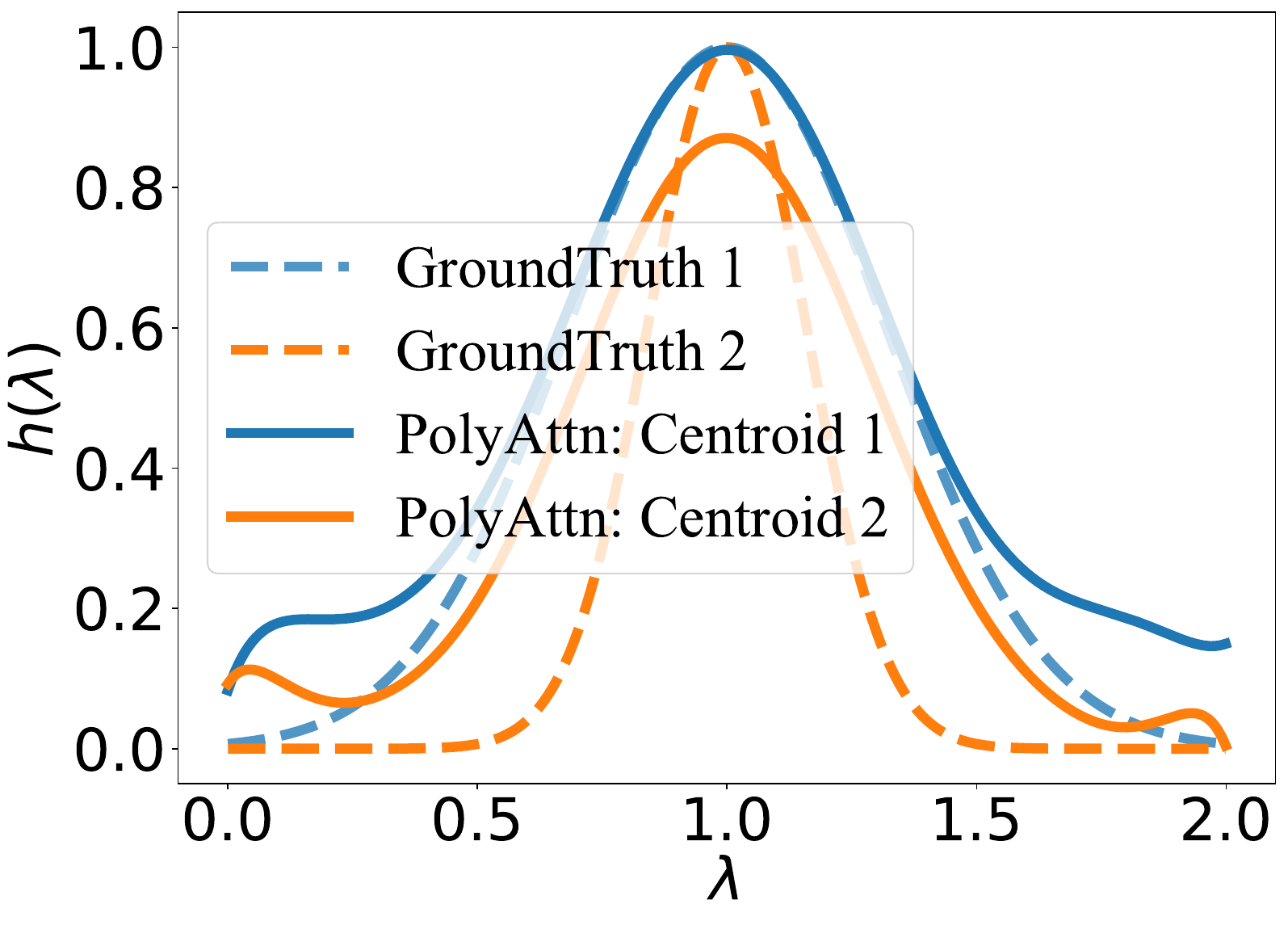}
  \caption{Mixed band-pass}
  \label{fig:image4}
\end{subfigure}

\caption{Learned filters of PolyAttn (Cheb).}
\label{filter_SBM_PolyAttn_fig}
% \vspace{-4mm}
\end{figure*}

\section{Experiments}
In this section, we conduct comprehensive experiments to evaluate the performance of the proposed PolyAttn and PolyFormer. Specifically, we first evaluate PolyAttn's ability on node-wise filtering using both synthetic and real-world datasets. 
Then, we execute node classification tasks on both small and large graphs to evaluate the performance of PolyFormer. 
We also conduct complexity and ablation comparison experiments.

\subsection{PolyAttn Experiments}

\subsubsection{Fitting Signals on Synthetic Datasets.}\label{FittingSignalsonSyntheticDatasets}

In this subsection, we evaluate the efficacy of PolyAttn as a node-wise filter on synthetic datasets. This evaluation highlights the enhanced capabilities of PolyAttn in learning individual filter patterns for each node, without the need for prior knowledge of predefined filters.

\textbf{Synthetic Datasets.} We use images with a resolution of $100 \times 100$ from the Image Processing in Matlab library\footnote{\href{https://ww2.mathworks.cn/products/image.html}{https://ww2.mathworks.cn/products/image.html}}. Each image can be represented as a 2D regular 4-neighborhood grid graph. The pixel values, ranging from $0$ to $1$, serve as node signals. For the $m$-th image, there exists an adjacency matrix $\mathbf{A}_m \in \mathbb{R}^{10000 \times 10000}$ and a node signal $\boldsymbol{x}_m \in \mathbb{R}^{10000}$. Based on the raw signal of each node, we apply two hybrid predefined filters to each image, as detailed in Table \ref{predefined_filter}. Models are expected to learn these predefined filtering patterns. More details can be seen in Appendix \ref{exp_polyatnn_sbm}.

\begin{table}[h!]
\vspace{-2mm}
\centering
\caption{Predefined filters on graph signals.}
% \vspace{-3mm}
\resizebox{\linewidth}{!}{
\begin{tabular}{lll}
\toprule
\textbf{Filters} & $\boldsymbol{h_1(\lambda)}$ & $\boldsymbol{h_2(\lambda)}$ \\
\midrule
Mixed low-pass & $h(\lambda) = e^{-5\lambda^2}$ & $h(\lambda) = e^{-20\lambda^2}$ \\
Mixed high-pass & $h(\lambda) = 1 - e^{-5\lambda^2}$ & $h(\lambda) = 1 - e^{-20\lambda^2}$ \\
Mixed band-pass & $h(\lambda) = e^{-5(\lambda-1)^2}$ & $h(\lambda) = e^{-20(\lambda-1)^2}$ \\
Mixed rejection-pass & $h(\lambda) = 1 - e^{-5(\lambda-1)^2}$ & $h(\lambda) = 1 - e^{-20(\lambda-1)^2}$ \\
% Low \& high-pass & $h(\lambda) = 1 - e^{-10\lambda^2}$ & $h(\lambda) = e^{-10\lambda^2}$ \\
% Band \& rejection-pass & $h(\lambda) = 1 - e^{-10(\lambda-1)^2}$ & $h(\lambda) = e^{-10(\lambda-1)^2}$ \\
Low \& high-pass & $h(\lambda) = e^{-10\lambda^2}$ & $h(\lambda) = 1 - e^{-10\lambda^2}$ \\
Band \& rejection-pass & $h(\lambda) = e^{-10(\lambda-1)^2}$  & $h(\lambda) = 1 - e^{-10(\lambda-1)^2}$  \\

\bottomrule
\end{tabular}
}
\label{predefined_filter}
\vspace{-2mm}
\end{table}

\textbf{Setup.} We compare PolyAttn with 5 baseline methods, 
%For the baseline, we use common GNN models, 
including GCN ~\citep{gcn_kipf2016semi}, GAT \citep{gat}, GPRGNN \cite{chien2021GPR-GNN}, BernNet \citep{bernnet}, and ChebNetII \citep{chebnetii_he2022convolutional}. To ensure a fair comparison, all models are constrained to one single layer and have approximately 5k parameters. The learning rate is uniformly set to $0.001$, the training epochs to $50,000$, and the early stopping threshold to $400$ iterations. We employ two metrics to evaluate each method: the sum of squared errors and the $R^2$ score.

\textbf{Results.} 
As demonstrated in Table \ref{result_SBM_PolyAtttn}, PolyAttn outperforms all baselines on all datasets. Compared to traditional polynomial GNNs, which employ unified coefficients for all nodes, PolyAttn uses tailored attention mechanisms for polynomial tokens to enable node-wise filtering. This design choice endows PolyAttn with greater expressive power. 
%Further evidence of this capability is provided in Figures \ref{polyattn:fig:image1} and \ref{polyattn:fig:image2}. 
Further evidence of this capability is provided in Figures \ref{filter_SBM_PolyAttn_fig}.
In the figure, filters learned on each node are divided into one of two clusters using the $k$-means \citep{kmeans_jain1988algorithms} algorithm, and the representative filter (centroid) for each cluster is plotted. PolyAttn is shown to successfully derive individual filter patterns without requiring any prior knowledge of predefined filters. This underscores PolyAttn's ability to learn graph filters for each node adaptively.

\begin{table}[htbp]
  % \vspace{-2mm}
  \small
  \centering
   \caption{Performance of PolyAttn on real-world datasets.}
   %\resizebox{0.5\textwidth}{!}{%
   \vspace{-2mm}
   \resizebox{\linewidth}{!}{
    \begin{tabular}{@{}lcccc@{}}
    \toprule
    & {\textbf{CS}} & {\textbf{Pubmed}} & {\textbf{Roman-empire}} & {\textbf{Questions}} \\
    \midrule
    UniFilter (Mono) & 95.32$\pm$0.24 & 89.61$\pm$0.44 & 73.44$\pm$0.80 & 73.19$\pm$1.52 \\
    PolyAttn (Mono) & 95.99$\pm$0.07 & 90.85$\pm$0.31 & 74.17$\pm$0.59 & 76.83$\pm$0.79 \\
    Improvement (\%) & 0.70 & 1.38 & 0.99 & 4.96 \\
    \midrule
    UniFilter (Bern) & 96.03$\pm$0.12 & 88.55$\pm$0.43 & 73.32$\pm$0.37 & 74.30$\pm$0.80 \\
    PolyAttn (Bern) & 95.84$\pm$0.21 & 90.18$\pm$0.41 & 76.33$\pm$0.30 & 77.79$\pm$0.74 \\
    Improvement (\%) & -0.20 & 1.80 & 4.11 & 4.70 \\
    \midrule

    UniFilter (Opt) & 95.08$\pm$0.23 & 89.61$\pm$0.32 & 76.33$\pm$0.37 & 75.38$\pm$0.86 \\
    PolyAttn (Opt) & 95.48$\pm$0.13 & 89.89$\pm$0.53 & 74.70$\pm$0.67 & 76.79$\pm$0.75 \\
    Improvement (\%) & 0.42 & 0.31 & -2.10 & 1.87 \\
    \midrule
    UniFilter (Cheb) & 96.17$\pm$0.10 & 88.65$\pm$0.35 & 72.81$\pm$0.73 & 74.55$\pm$0.78 \\
    PolyAttn (Cheb) & 96.03$\pm$0.15 & 89.85$\pm$0.46 & 74.03$\pm$0.45 & 75.90$\pm$0.72 \\
    Improvement (\%) & -0.15 & 1.35 & 1.68 & 1.81 \\
    \bottomrule
    \end{tabular}%
    }
  \label{result_real_PolyAttn}%
  \vspace{-2mm}
\end{table}%

\begin{table*}[thbp]
  \centering
   \caption{Performance of PolyFormer on node classification. ``OOM'' means ``out of memory,'' and ``*'' indicates the use of truncated eigenvalues and eigenvectors as suggested by ~\cite{specformer}.}

    \vspace{-2mm}
    \resizebox{\textwidth}{!}{
    \begin{tabular}{ccccc|cccccc}
    \toprule
          & \multicolumn{4}{c|}{\textbf{Homophilic}} & \multicolumn{6}{c}{\textbf{Heterophilic}} \\
    \midrule
    Datasets      & \textbf{Cite.} & \textbf{CS}    & \textbf{Pubm.} & \textbf{Phys.} & \textbf{Cham.} & \textbf{Squi.} & \textbf{Mine.} & \textbf{Tolo.} & \textbf{Roman.} & \textbf{Ques.} \\
    % $|V|$   & 3,327 & 18,333 & 19,717 & 34,493 & 890 & 2,223 & 10,000 & 11,758 & 22,662 & 48,921 \\
    % $|E|$   & 9,104 & 163,788 & 44,324 & 495,924  & 17,708 & 93,996 & 39,402 & 519,000 & 32,927 & 153,540 \\

    {Nodes} & 3,327 & 18,333 & 19,717 & 34,493 & 890 & 2,223 & 10,000 & 11,758 & 22,662 & 48,921 \\
    {Edges} & 9,104 & 163,788 & 44,324 & 495,924 & 17,708 & 93,996 & 39,402 & 519,000 & 32,927 & 153,540 \\
    {Features} & 3,703 & 6,805 & 500 & 8,415 & 2,325 & 2,089 & 7 & 10 & 300 & 301 \\
    {Classes} & 6 & 15 & 3 & 5 & 5 & 5 & 2 & 2 & 18 & 2 \\
    % \midrule
    % \begin{tabular}{ccccc|cccc}
    % \toprule
    %       & \multicolumn{4}{c|}{\textbf{Homophilic}} & \multicolumn{4}{c}{\textbf{Heterophilic}} \\
    % \midrule
          % & \textbf{Cite.} & \textbf{CS}    & \textbf{Pub.} & \textbf{Phy.} & \textbf{Cham.} & \textbf{Squi.} & \textbf{Mine.} & \textbf{Tolokers} \\
    \midrule
    MLP &78.74$_{\pm\text{0.64}}$ &95.53$_{\pm\text{0.13}}$ &87.06$_{\pm\text{0.35}}$ &97.10$_{\pm\text{0.71}}$ &41.84$_{\pm\text{1.81}}$ &39.19$_{\pm\text{1.81}}$ &50.97$_{\pm\text{0.54}}$ &74.12$_{\pm\text{0.48}}$ &66.64$_{\pm\text{0.32}}$ &71.87$_{\pm\text{0.41}}$\\

    GCN &80.16$_{\pm\text{1.09}}$ &94.95$_{\pm\text{0.17}}$ &87.34$_{\pm\text{0.37}}$ &97.74$_{\pm\text{0.35}}$ &43.43$_{\pm\text{1.92}}$ &41.30$_{\pm\text{0.94}}$ &72.23$_{\pm\text{0.56}}$ &77.22$_{\pm\text{0.73}}$ &53.45$_{\pm\text{0.27}}$ &76.28$_{\pm\text{0.64}}$\\

    GAT &80.67$_{\pm\text{1.05}}$ &93.93$_{\pm\text{0.26}}$ &86.55$_{\pm\text{0.36}}$ &97.82$_{\pm\text{0.28}}$ &40.14$_{\pm\text{1.57}}$ &35.09$_{\pm\text{0.70}}$&81.39$_{\pm\text{1.69}}$ &77.87$_{\pm\text{1.00}}$ &51.51$_{\pm\text{0.86}}$ &74.94$_{\pm\text{0.56}}$\\

    GPRGNN &80.61$_{\pm\text{0.75}}$ &95.26$_{\pm\text{0.15}}$ &\cemp91.00$_{\pm\text{0.34}}$ &97.74$_{\pm\text{0.35}}$ &42.28$_{\pm\text{2.87}}$ &41.09$_{\pm\text{1.18}}$ &90.10$_{\pm\text{0.34}}$ &77.25$_{\pm\text{0.61}}$ &74.08$_{\pm\text{0.54}}$ &74.36$_{\pm\text{0.67}}$\\

    BernNet &79.63$_{\pm\text{0.78}}$ &95.42$_{\pm\text{0.29}}$ &90.56$_{\pm\text{0.40}}$ &97.64$_{\pm\text{0.38}}$ &42.57$_{\pm\text{2.72}}$ &39.30$_{\pm\text{1.37}}$ &77.93$_{\pm\text{0.59}}$ &76.83$_{\pm\text{0.53}}$ &72.70$_{\pm\text{0.30}}$ &74.25$_{\pm\text{0.73}}$\\
    %BernNet & 79.63±0.78 & 95.42±0.29 & 90.56±0.40 & 97.64±0.38 & 77.93±0.59 & 76.83±0.53 & 72.70±0.30 & 74.25±0.73 \\
    
    ChebNetII &80.25$_{\pm\text{0.65}}$ & \cemp96.33$_{\pm\text{0.12}}$ &90.60$_{\pm\text{0.17}}$ &97.25$_{\pm\text{0.78}}$ &42.67$_{\pm\text{1.43}}$ &41.22$_{\pm\text{0.37}}$ &83.64$_{\pm\text{0.40}}$ &79.23$_{\pm\text{0.43}}$ &74.64$_{\pm\text{0.39}}$ &74.41$_{\pm\text{0.58}}$\\

    OptBasisGNN& 80.58$_{\pm\text{0.82}}$ & 94.77$_{\pm\text{0.23}}$ & 90.30$_{\pm\text{0.23}}$ & 97.64$_{\pm\text{0.48}}$ & 41.23$_{\pm\text{3.16}}$ & 42.34$_{\pm\text{2.74}}$ & 89.74$_{\pm\text{1.03}}$ & 81.08$_{\pm\text{0.96}}$ & 76.91$_{\pm\text{0.37}}$ & 73.82$_{\pm\text{0.83}}$ \\

    \midrule
    
    DSF-GPR-R & 78.22$_{\pm\text{0.29}}$ & 96.25$_{\pm\text{0.12}}$ & 90.51$_{\pm\text{0.07}}$ & 98.07$_{\pm\text{0.36}}$ & 43.82$_{\pm\text{1.51}}$ & 41.31$_{\pm\text{1.07}}$ & 89.51$_{\pm\text{0.00}}$ & 79.74$_{\pm\text{1.19}}$ & 75.18$_{\pm\text{0.37}}$ & 74.16$_{\pm\text{1.07}}$ \\

    DSF-Bern-R & 78.27$_{\pm\text{0.26}}$ & 96.28$_{\pm\text{0.09}}$ & 90.52$_{\pm\text{0.10}}$ & \cemp98.47$_{\pm\text{0.10}}$ & 44.07$_{\pm\text{2.20}}$ & 39.69$_{\pm\text{1.56}}$ & 77.18$_{\pm\text{0.05}}$ & 75.78$_{\pm\text{0.98}}$ & 75.39$_{\pm\text{0.30}}$ & 73.81$_{\pm\text{0.39}}$ \\
    
    \midrule

    Transformer &78.70$_{\pm\text{0.59}}$ &OOM &89.10$_{\pm\text{0.43}}$ &OOM  &43.27$_{\pm\text{1.65}}$ &39.82$_{\pm\text{0.84}}$&50.29$_{\pm\text{1.09}}$ &74.24$_{\pm\text{0.58}}$ &65.29$_{\pm\text{0.47}}$ &OOM\\
    %Transformer & 78.70±0.59 & OOM   & 89.10±0.43 & OOM   & 50.29±1.09 & 74.24±0.58 & 65.29±0.47 & OOM \\
    
    Specformer &\cemp81.69$_{\pm\text{0.78}}$ &96.07$_{\pm\text{0.10}}$ &89.94$_{\pm\text{0.33}}$ &97.70$_{\pm\text{0.60}}$* &42.82$_{\pm\text{2.54}}$ &40.20$_{\pm\text{0.53}}$ &89.93$_{\pm\text{0.41}}$ &80.42$_{\pm\text{0.55}}$ &69.94$_{\pm\text{0.34}}$ &76.49$_{\pm\text{0.58}}$*\\
    %Specformer & 81.69±0.78 & 96.07±0.10 & 89.94±0.33 & 97.70±0.60* & 89.93±0.41 & 80.42±0.55 & 69.94±0.34 & 76.49±0.58* \\
    
    NAGphormer &79.77$_{\pm\text{0.81}}$ &95.89$_{\pm\text{0.13}}$ &89.65$_{\pm\text{0.45}}$ &97.23$_{\pm\text{0.23}}$ &40.36$_{\pm\text{1.77}}$ &39.79$_{\pm\text{0.84}}$ &88.06$_{\pm\text{0.43}}$ &81.57$_{\pm\text{0.44}}$ &74.45$_{\pm\text{0.48}}$ &75.13$_{\pm\text{0.70}}$ \\

    GOAT & 76.40$_{\pm\text{0.43}}$ & 95.12$_{\pm\text{0.21}}$ & 90.63$_{\pm\text{0.26}}$ & 97.29$_{\pm\text{0.24}}$ & 41.55$_{\pm\text{1.20}}$ & 38.71$_{\pm\text{0.56}}$ & 82.90$_{\pm\text{0.62}}$ & 83.13$_{\pm\text{1.19}}$ & 72.30$_{\pm\text{0.48}}$ & 75.95$_{\pm\text{1.38}}$ \\
    
    NodeFormer & 80.35$_{\pm\text{0.75}}$ & 95.64$_{\pm\text{0.23}}$ & \cemp91.20$_{\pm\text{0.36}}$ & 96.45$_{\pm\text{0.28}}$ & 43.73$_{\pm\text{3.26}}$ & 37.07$_{\pm\text{9.16}}$ & 86.91$_{\pm\text{1.02}}$ & 78.34$_{\pm\text{0.98}}$ & 74.29$_{\pm\text{0.75}}$ & 74.48$_{\pm\text{1.32}}$ \\

    SGformer & 81.11$_{\pm\text{1.08}}$ & 94.86$_{\pm\text{0.38}}$ & 89.57$_{\pm\text{0.90}}$ & 97.96$_{\pm\text{0.81}}$ & 44.21$_{\pm\text{3.06}}$ & \cemp43.74$_{\pm\text{2.51}}$ & 77.69$_{\pm\text{0.96}}$ & 82.07$_{\pm\text{1.18}}$ & 73.91$_{\pm\text{0.79}}$ & 77.06$_{\pm\text{1.20}}$ \\

    \midrule

    PolyFormer (Opt) & 79.95$_{\pm\text{0.61}}$ & 95.87$_{\pm\text{0.23}}$ & 90.09$_{\pm\text{0.36}}$ & 97.66$_{\pm\text{0.14}}$ & \cemp47.55$_{\pm\text{2.61}}$ & \cemp43.86$_{\pm\text{1.46}}$ & \cemp91.93$_{\pm\text{0.37}}$ & \cemp83.15$_{\pm\text{0.49}}$ & \cemp77.15$_{\pm\text{0.33}}$ & \cemp77.69$_{\pm\text{0.92}}$ \\
    
    PolyFormer (Mono) &\cemp82.37$_{\pm\text{0.65}}$ &\cemp96.49$_{\pm\text{0.09}}$ &\cemp91.01$_{\pm\text{0.41}}$ &\cemp98.42$_{\pm\text{0.16}}$ & \cemp46.86$_{\pm\text{1.61}}$ & \cemp42.56$_{\pm\text{0.96}}$ &\cemp90.69$_{\pm\text{0.38}}$ &\cemp84.00$_{\pm\text{0.45}}$ &\cemp78.89$_{\pm\text{0.39}}$ &\cemp77.46$_{\pm\text{0.65}}$\\

    PolyFormer (Bern) &\cemp81.39$_{\pm\text{0.61}}$ & \cemp96.34$_{\pm\text{0.15}}$ & \cemp91.31$_{\pm\text{0.35}}$ & \cemp98.34$_{\pm\text{0.23}}$ & \cemp46.99$_{\pm\text{2.39}}$ & \cemp44.86$_{\pm\text{0.98}}$ & \cemp92.02$_{\pm\text{0.32}}$ & \cemp84.32$_{\pm\text{0.59}}$ & \cemp77.64$_{\pm\text{0.33}}$ & \cemp78.32$_{\pm\text{0.67}}$ \\
    
    PolyFormer (Cheb) &\cemp81.80$_{\pm\text{0.76}}$ &\cemp96.49$_{\pm\text{0.17}}$ &90.68$_{\pm\text{0.31}}$ &\cemp98.08$_{\pm\text{0.27}}$ & \cemp45.35$_{\pm\text{2.97}}$ & 41.83$_{\pm\text{1.18}}$ & \cemp91.90$_{\pm\text{0.35}}$ &\cemp83.88$_{\pm\text{0.33}}$ &\cemp80.27$_{\pm\text{0.39}}$ &\cemp77.26$_{\pm\text{0.50}}$\\
    \bottomrule
    \end{tabular}%
    }
   \label{PolyFormer_realdataset}%
   \vspace{-2mm}
\end{table*}%

% 牺牲掉的图3
\begin{figure}[htbp]
\centering
\begin{subfigure}{0.23\textwidth}
  \centering
  \includegraphics[width=\linewidth]{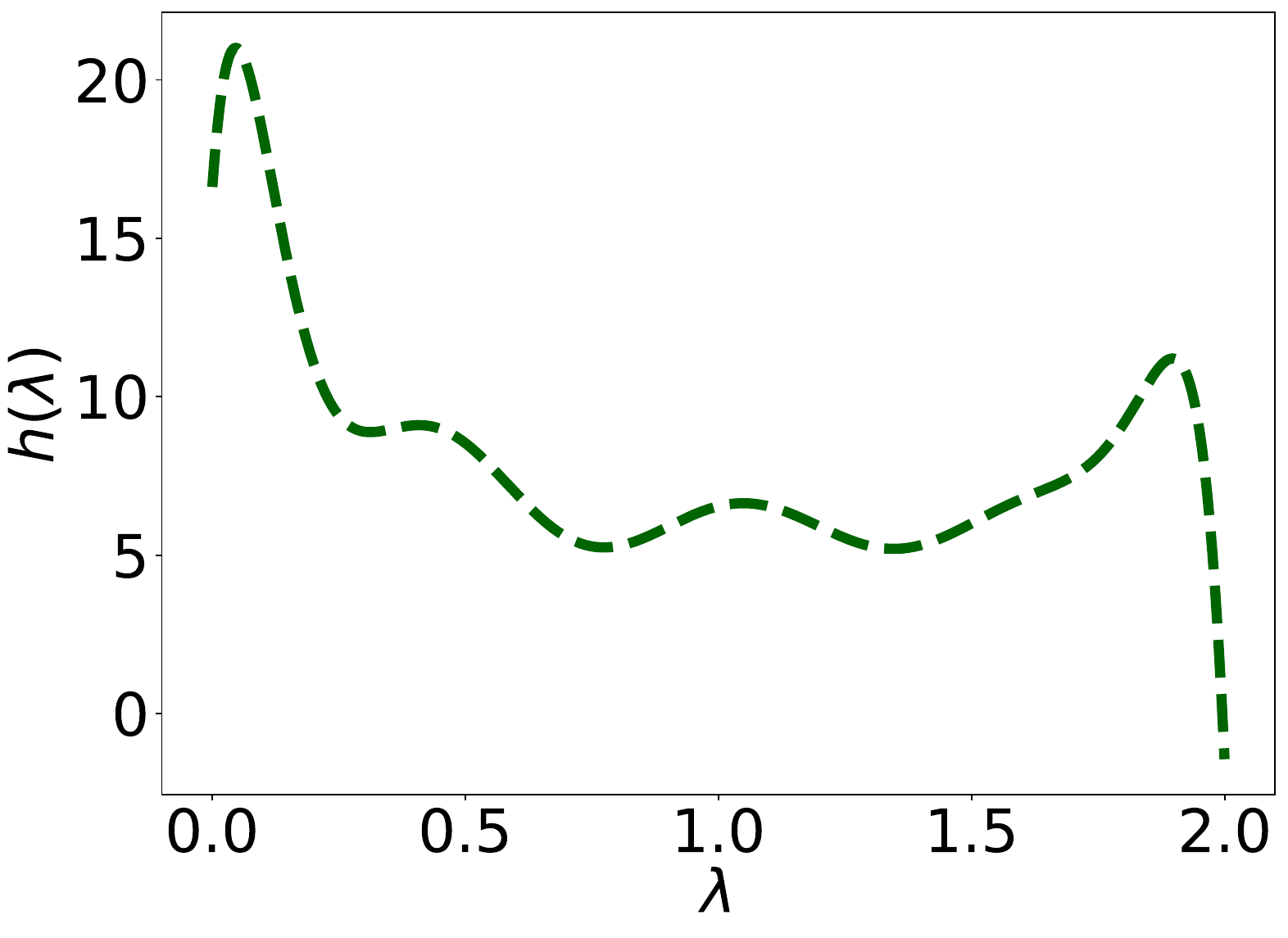}
  % \caption{Filter of UniFilter}
  \caption{UniFilter on Pubmed}
  \label{realfig:image1}
\end{subfigure}%
\begin{subfigure}{0.23\textwidth}
  \centering
  \includegraphics[width=\linewidth]{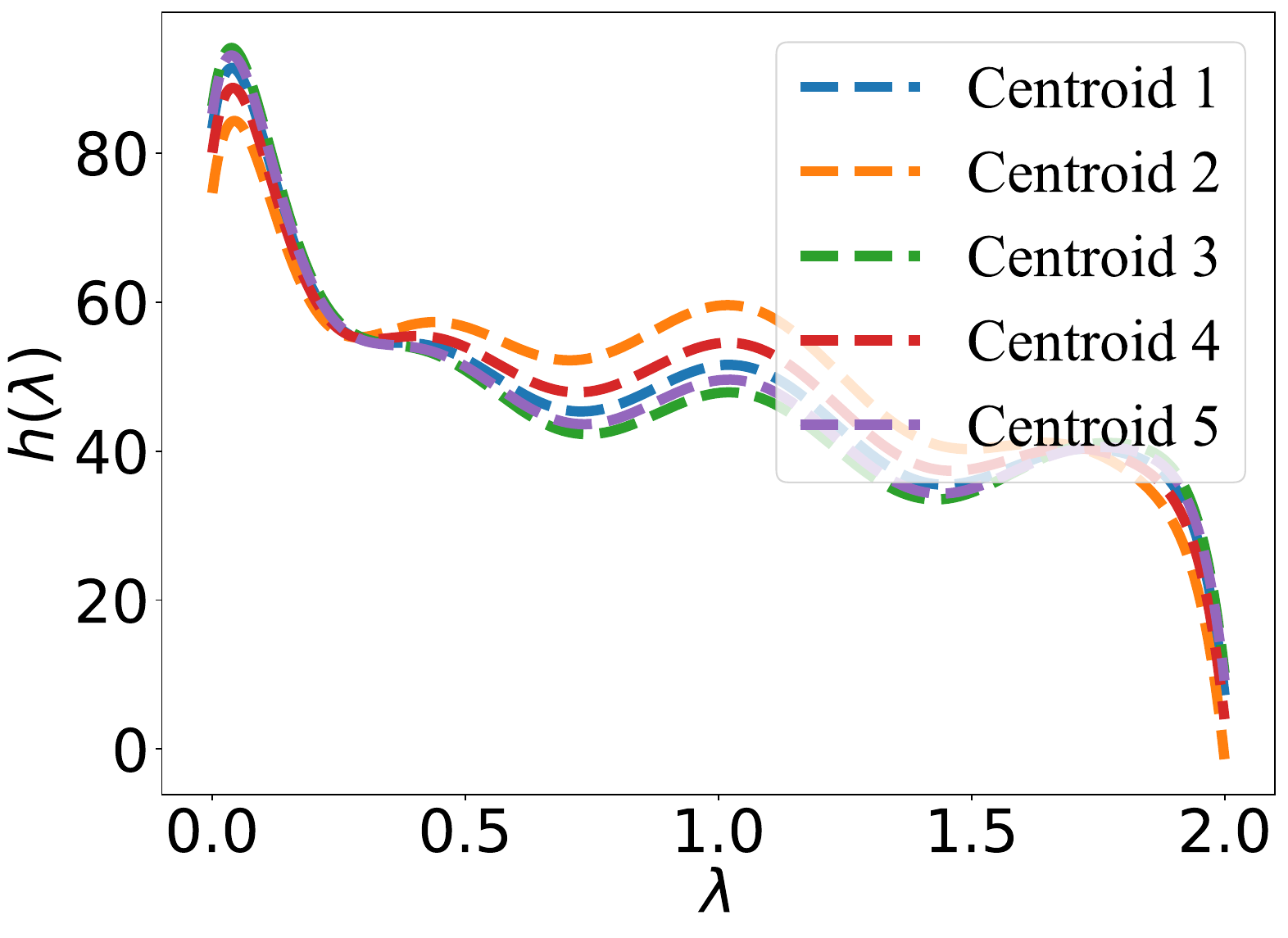}
  % \caption{Filters of PolyAttn}
  \caption{PolyAttn on Pubmed}
  \label{realfig:image2}
\end{subfigure}%
\\
\begin{subfigure}{0.22\textwidth}
  \centering
  \includegraphics[width=\linewidth]{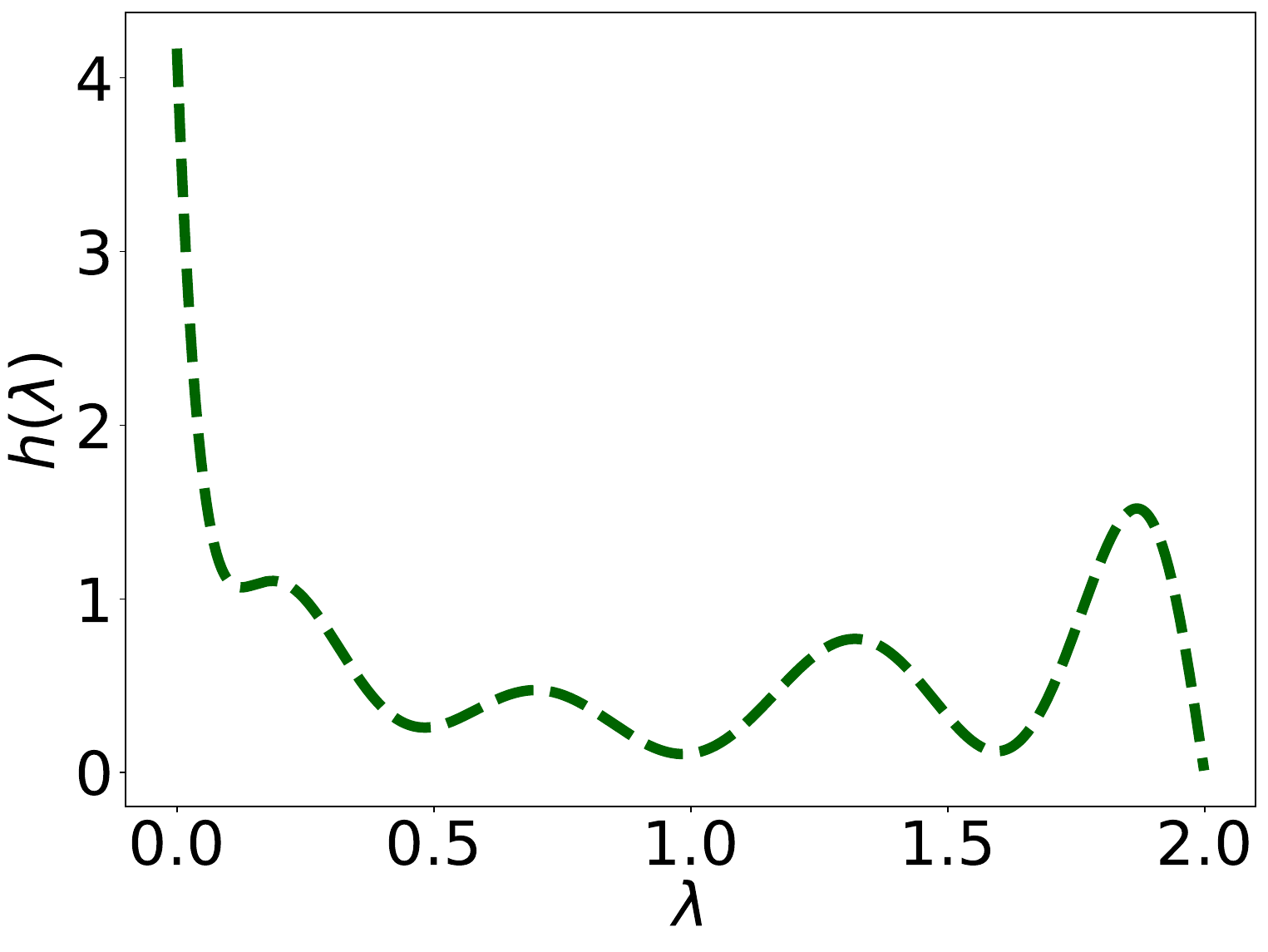}
  % \caption{Filter of UniFilter}
  \caption{UniFilter on Questions}
  \label{realfig:image3}
\end{subfigure}%
\begin{subfigure}{0.238\textwidth}
  \centering
  \includegraphics[width=\linewidth]{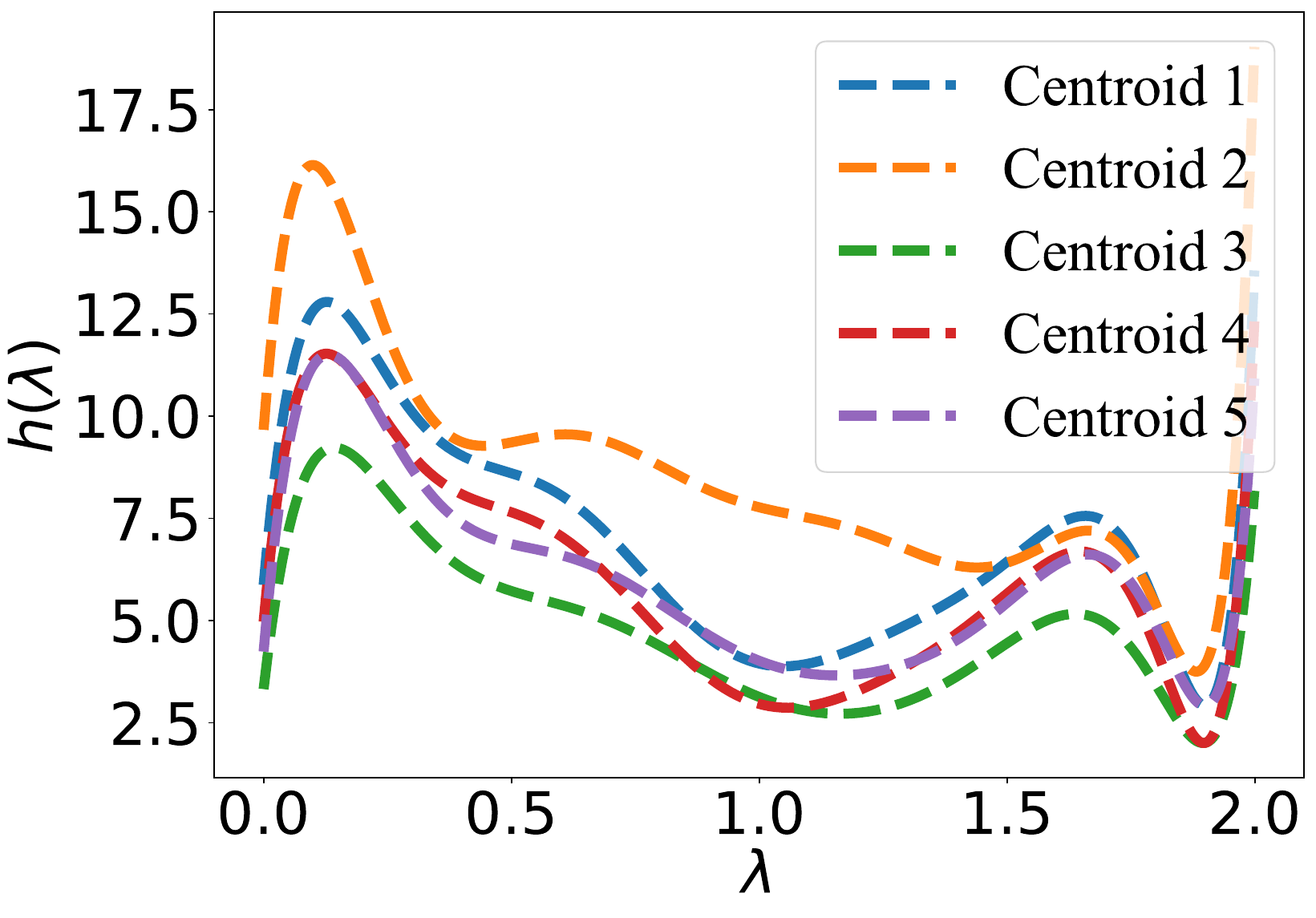}
  % \caption{Filters of PolyAttn}
  \caption{PolyAttn on Questions}
  \label{realfig:image4}
\end{subfigure}%
% \caption{Filters Learned by UniFilter (Left) and PolyAttn (Right) on Homophilic Graph Pubmed (Top) and Heterophilic Graph Questions (Bottom).}
\caption{Filters learned by UniFilter (left) and PolyAttn (right) on the homophilic graph Pubmed (top) and the heterophilic graph Questions (bottom).}
% \caption{Filters Learned by UniFilter (left) and PolyAttn (right) on Homophilic Graphs (Top: PubMed) and Heterophilic Graphs (Bottom: Questions).}
\label{filter_real_PolyAttn_fig}
\vspace{-2mm}
\end{figure}

\subsubsection{Performance on Real-world Datasets} \label{PolyAttn_on_realdata}

We evaluate the efficacy of PolyAttn as a node-wise filter on real-world datasets.

% In this subsection, the efficacy of PolyAttn, when employed as a node-wise filter, is evaluated using real-world datasets.

\textbf{Setup.}
For homophilic datasets CS and Pubmed, we employ a random split: 60\% for the training set, 20\% for the validation set, and 20\% for the test set, following the approach of \citep{bernnet}. For heterophilic graphs Roman-empire and Questions, we adhere to the partitioning scheme provided in \citep{dataset_russianhetegraph}, allocating 50\% of the data for training, 25\% for validation, and 25\% for testing.
% The datasets for homophilic graphs are divided into 60\% training, 20\% validation, and 20\% testing sets, while those for heterophilic graphs follow a 50\%, 25\%, 25\% split. 
We employ node-unified filters, which learn shared polynomial coefficients for all nodes, denoted as ``UniFilter''. Correspondingly, we use node-wise filters based on PolyAttn. All models are configured with a single filtering layer and the same truncated order to ensure a fair comparison. More details are listed in Appendix~\ref{exp_polyattn_real}.

\textbf{Results.}
Table \ref{result_real_PolyAttn} shows the mean accuracies with a 95\% confidence interval over 10 runs.
%As shown in Table \ref{result_real_PolyAttn}, 
We observe that PolyAttn performs better on both homophilic and heterophilic graphs, with especially notable improvements on the latter one, which suggests the benefits of its node-wise filtering ability. 
Further insights are illustrated in Figures \ref{filter_real_PolyAttn_fig}, which show the learned filters by ``UniFilter (Cheb)'' and ``PolyAttn (Cheb)'' on Pubmed and Questions. The node-wise filters learned by PolyAttn are categorized into one of five clusters using the k-means algorithm \citep{kmeans_jain1988algorithms}.
Interestingly, we observe that the filters learned by PolyAttn bear resemblance to the node-unified filter, yet display a greater level of sophistication. Given that PolyAttn achieves a relative improvement of up to 4.96\% over UniFilter, this suggests that node-wise filters are essential for enhancing expressiveness. 
% More illustrations can be found in Appendix \ref{filters_of_polyattn}.

\subsection{PolyFormer Experiments}

\subsubsection{Node Classification}

In this subsection, we evaluate the proposed PolyFormer on real-world datasets, which encompass both homophilic and heterophilic types, demonstrating the model's exceptional performance.

\textbf{Setup.}
We employ datasets including four homophilic datasets \citep{datasest_coraciteseer_sen2008collective, dataset_cs_shchur2018pitfalls} and six heterophilic datasets \citep{dataset_russianhetegraph}.
For homophilic datasets including Citeseer, CS, Pubmed, and Physics, we employ a random split: 60\% for the training set, 20\% for the validation set, and 20\% for the test set, following the approach of \citep{bernnet}.
For heterophilic graphs, specifically Minesweeper, Tolokers, Roman-empire, and Questions, we adhere to the partitioning scheme provided in \citep{dataset_russianhetegraph}, allocating 50\% of the data for training, 25\% for validation, and 25\% for testing. Similarly, for the processed heterophilic datasets Chameleon and Squirrel, we also utilize the splits specified in \citep{dataset_russianhetegraph}.
As for the baselines, we select several recent state-of-the-art spectral GNNs \cite{chien2021GPR-GNN, bernnet, chebnetii_he2022convolutional, optbasis_guo2023graph}, as well as extended node-wise filters \cite{dsf_guo2023graph}. Additionally, our comparison includes competitive Graph Transformer models \cite{nagphormer, specformer, goat_kong2023goat, nodeformer, SGFormer}.  More details are available in Appendix~\ref{exp_polyformer_common}.

\textbf{Results.}
As shown in Table \ref{PolyFormer_realdataset}, our model consistently outperforms most baseline models, especially excelling on heterophilic datasets.
% Notably, when compared with spectral GNNs like ChebNetII, which utilize advanced techniques like Chebyshev Interpolation or node-wise filters DSF based on the additional positional encoding, our model showcases superior performance. Such results suggest that the introduction of node-wise filters PolyAttn significantly boosts our model's expressive power.
Notably, when compared with spectral GNNs such as ChebNetII, which utilize advanced techniques like Chebyshev Interpolation, and node-wise filters like DSF based on additional positional encoding, our model showcases superior performance. These results suggest that the introduction of node-wise filters, PolyAttn, significantly boosts the expressive power of our model.
Furthermore, our model maintains competitive performance against transformer-based approaches. This observation indicates that focusing on information within a limited scope, i.e., a truncation of polynomial basis, provides the necessary expressiveness for achieving competitive results. Conversely, taking all node pairs into account may introduce redundant noise that diminishes the model's performance.

\subsubsection{Node Classifications on Large-scale Datasets}

In this subsection, we extend our model to large-scale datasets, demonstrating the scalability of our model, facilitated by its efficient node tokenization methods and scalable node-wise filter.

\textbf{Setup.}
We perform node classification tasks on two expansive citation networks: ogbn-arxiv and ogbn-papers100M \citep{ogb}, in addition to two large-scale heterophilic graphs: Twitch-Gamers and Pokec, sourced from \citep{linkx} to demonstrating the scalability of our model. 
% More information is provided in Appendix \ref{data_description}. 
For the citation datasets, experiments are conducted using the given data splits in \citep{ogb}. For the datasets Twitch-gamers and Pokec, we utilize the five fixed data splits provided in \citep{linkx}. 
% For large-scale datasets, we utilize the split from \citep{linkx} for Twitch-gamers and pokec. Meanwhile, for ogbn-arxiv and ogbn-papers100M, we adhere to the given splits as presented in \citep{ogb}.
We select common GNN models, including ~\citep{gcn_kipf2016semi, chien2021GPR-GNN, Chebnet}. For Graph Transformer models, we use expressive Specformer ~\citep{specformer}, and three scalable baselines NAGphormer ~\citep{nagphormer}, Nodeformer ~\citep{nodeformer} and SGFormer \citep{SGFormer}. More details are available in Appendix~\ref{exp_polyformer_large}. %The mean test accuracy of multiple runs is presented.

% \{Baseline.}

\textbf{Results.}
Table~\ref{PolyFormer_largedataset} shows the mean accuracy over multiple runs.
Due to our efficient node tokenization techniques and scalable node-wise filters, PolyFormer exhibits great scalability up to the graph ogbn-papers100M, which has over \textbf{100 million nodes}. In contrast, models such as NAGphormer \citep{nagphormer} and Specformer \citep{specformer} rely on Laplacian eigenvectors or eigenvalues, which constrains their scalability. Moreover, by leveraging expressive PolyAttn, our model exhibits superior performance.

% \cite{citation_key}.
% ``-" means ``out of memory" or failing to finish preprocessing in 24h.

\begin{table}[htbp]
% \vspace{-2mm}
  \small
  \centering
    % \caption{Performance of PolyFormer on Node Classification Tasks on Large-Scale Dataset.}
    \caption{Performance of PolyFormer for node classification on large-scale datasets. ``-'' means ``out of memory'' or failing to complete preprocessing within 24 hours.}
    % \vspace{-2mm}
  \resizebox{\linewidth}{!}{%
  % resizebox{\linewidth}{!}{
    \begin{tabular}{ccccc}
    \toprule
    Datasets         & \textbf{Twitch} & \textbf{arxiv} & \textbf{Pokec} & \textbf{papers100M} \\
    % $\|V\|$  & 168,114 & 169,343 & 1,632,803 & 111,059,956 \\
    % $\|E\|$  & 6,797,557 & 1,166,243 & 30,622,564 & 1,615,685,872 \\
    % $H(G)$  & 0.545 & 0.66  & 0.445 & - \\
    {Nodes} & 168,114 & 169,343 & 1,632,803 & 111,059,956 \\
    {Edges} & 6,797,557 & 1,166,243 & 30,622,564 & 1,615,685,872 \\
    {Features} & 7 & 128 & 65 & 128 \\
    {Classes} & 2 & 40 & 2 & 172 \\
    \midrule
    MLP   & 60.92±0.07 & 55.50±0.23 & 62.37±0.02 & 47.24±0.31 \\
    GCN   & 62.18±0.26 & 71.74±0.29 & 75.45±0.17 & - \\
    ChebNet & 62.31±0.37 & 71.12±0.22 & -     & - \\
    GPR-GNN & 62.59±0.38 & 71.78±0.18 & \underline{80.74±0.22} & 65.89±0.35 \\
    % \midrule
    Specformer & 64.22±0.04 & 72.37±0.18 & -     & - \\
    NAGphormer & 64.38±0.04 & 71.04±0.94 & -     & - \\
    NodeFormer & 61.12±0.05 & 59.90±0.42 & 70.32±0.45 & - \\
    SGFormer & \textbf{65.26±0.26} & \textbf{72.63±0.13} & {73.76±0.24} & \underline{66.01±0.37} \\
    \midrule
    PolyFormer & \underline{64.79±0.10} & \underline{72.42±0.19} & \textbf{82.29±0.14} & \textbf{67.11±0.20} \\
    \bottomrule
    \end{tabular}
    }
  \label{PolyFormer_largedataset}%
  \vspace{-3mm}
\end{table}%

\subsection{Complexity Comparison}
In this subsection, we evaluate PolyFormer in comparison to other attention-based models concerning accuracy, time, and GPU memory consumption.

\begin{figure}[htbp]
\vspace{-1mm}
\centering
\resizebox{0.48\textwidth}{!}{%
\includegraphics[width=1.0\textwidth]{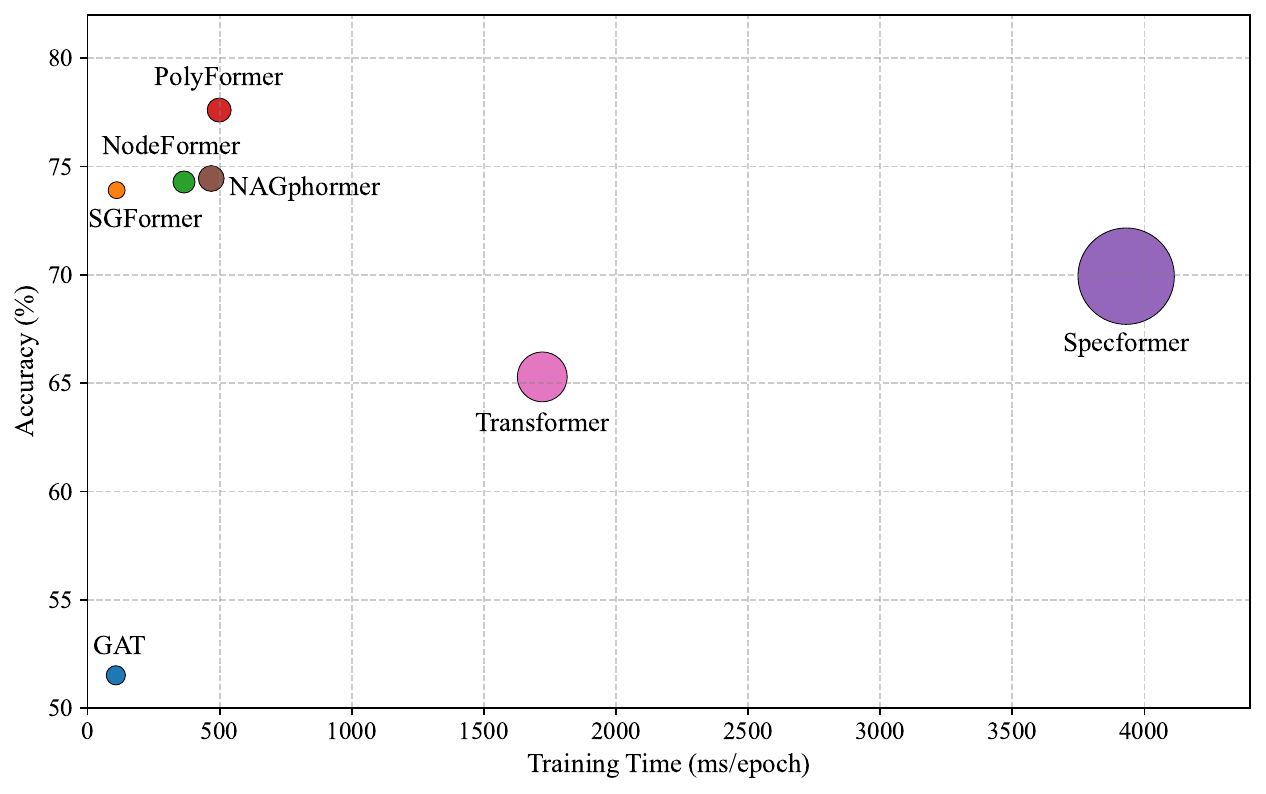}
}
% \caption{Accuracy, train time, and maximum GPU memory consumption comparison. In the Figure, the x-axis and y-axis denote accuracy and train time respectively, while the area of the circle implies the maximum GPU memory consumption.}
% \vspace{-3mm}
\caption{Accuracy, training time, and relative maximum GPU memory consumption comparison on Roman-empire.}
\label{time&mem}
\end{figure}

\textbf{Results.}
As shown in Figure \ref{time&mem}, the x-axis represents training time, and the y-axis represents accuracy, while the area of the circle indicates the relative maximum GPU memory consumption. PolyFormer achieves the best performance with low training time and GPU memory overhead. With the optimal accuracy obtained, the model possesses slightly more parameters, leading to a marginally longer training time, which is still acceptable in practice. 
% Additionally, we record the processing times for several methods: approximately 1110 seconds for Specformer, 40 seconds for NAGphormer, and only 0.36 seconds for PolyFormer. Given the significant reduction in preprocessing time, PolyFormer proves highly efficient in practical applications. Furthermore, PolyFormer's capability for mini-batch training can result in even lower GPU consumption than reported, thus facilitating scalability to graphs of any size.
Additionally, we record the processing times for several methods: approximately 1110 seconds for Laplacian eigendecomposition in Specformer, 40 seconds for truncated Laplacian eigendecomposition in NAGphormer, and merely 0.36 seconds for calculating polynomial tokens in PolyFormer. Given the significant reduction in preprocessing time, PolyFormer proves highly efficient in practical applications. Furthermore, PolyFormer's capability for mini-batch training can result in even lower GPU consumption than reported, thus facilitating scalability to graphs of any size.

\subsection{Ablation Comparison}

In this subsection, we present the results of the ablation study, which compares the proposed PolyAttn mechanism that uses the tanh activation function with the vanilla attention (self-attention) mechanism, which uses the softmax activation function. We compare the ability of these two mechanisms to fit the predefined node-wise filters when given the same polynomial tokens. For example, SelfAttn (Mono) and PolyAttn (Mono) respectively refer to self-attention and PolyAttn, with node tokens based on the Monomial basis. The other experimental settings are consistent with those described in \ref{FittingSignalsonSyntheticDatasets}.

\textbf{Results.} Table \ref{result_SBM_PolyAtttn_ab} presents the results of the ablation experiments, which show the sum of squared errors in fitting node-wise filters - where lower values indicate better fitting performance. It is clear that the proposed PolyAttn that uses the tanh activation function outperforms the vanilla attention mechanism in accurately fitting node-wise filters, indicating the validity of our proposed PolyAttn.

\begin{table}[t]
  \centering
% \caption{Performance of SelfAttn and PolyAttn on Synthetic Datasets, presented as the sum of squared errors (with values closer to 0 indicating higher accuracy).}
\caption{Performance comparison of SelfAttn and PolyAttn in fitting node-wise filters. Lower sum of squared errors indicates better performance.}
\vspace{-3mm}
  \resizebox{0.46\textwidth}{!}{%
    \begin{tabular}{lcccccc}
    \toprule
    \textbf{Model (5k para.)} & \textbf{Low-pass} & \textbf{High-pass} & \textbf{Band-pass} & \textbf{Rejection-pass} \\
    \midrule

    SelfAttn (Mono) & 0.4089 & 0.4037 & 2.1106 & 1.9176  \\
        
    {PolyAttn (Mono)} & {0.2550} & {0.2631} & 1.3798 & 1.4025  \\

    \midrule
    
    SelfAttn (Bern)  & 0.2758 & 0.2320 & 0.4295 & 0.4284  \\ 
    
    {PolyAttn (Bern)} & {0.0842} & 0.1120 & 0.2719 & 0.3337\\

    \midrule
    
    SelfAttn (Opt) & 0.2107 & 0.0249 & 0.1643 & 0.3026 \\
    
    {PolyAttn (Opt)} & 0.1922 & {0.0103} & {0.0701} & {0.2275} \\

    \midrule
    
    SelfAttn (Cheb) & 0.1226 & 0.2241 & 0.1674 & 0.3853  \\
    
    {PolyAttn (Cheb)} & {0.1467} & {0.0148} & {0.0782} & {0.1949}  \\

    \bottomrule
    \end{tabular}%
    }
  \label{result_SBM_PolyAtttn_ab}%
  \vspace{-3mm}
\end{table}%

\section{Conclusion}

In this study, we introduce PolyAttn, an attention-based node-wise filter. PolyAttn utilizes polynomial bases to capture spectral information efficiently, outperforming traditional node-unified filters in expressiveness while maintaining scalability and efficiency. Furthermore, we present PolyFormer, a scalable Graph Transformer tailored for node-level tasks.  PolyFormer strikes a balance between expressive power and scalability. Extensive empirical evaluations confirm the superior performance, efficiency, and scalability of PolyFormer. A promising future direction involves enhancing PolyAttn and PolyFormer through the incorporation of more advanced polynomial approximation and graph spectral techniques.

\begin{acks}
This research was supported in part by National Natural Science Foundation of China (No. U2241212, No. 61932001), by National Science and Technology Major Project (2022ZD0114800), by Beijing Natural Science Foundation (No. 4222028), by Beijing Outstanding Young Scientist Program No.BJJWZYJH012019100020098, and by Huawei-Renmin University joint program on Information Retrieval. We also wish to acknowledge the support provided by the fund for building world-class universities (disciplines) of Renmin University of China, by Engineering Research Center of Next-Generation Intelligent Search and Recommendation, Ministry of Education, Intelligent Social Governance Interdisciplinary Platform, Major Innovation \& Planning Interdisciplinary Platform for the “Double-First Class” Initiative, Public Policy and Decision-making Research Lab, and Public Computing Cloud, Renmin University of China.
\end{acks}

\clearpage

\bibliographystyle{ACM-Reference-Format}
\bibliography{main}

\appendix

\section{Notations}

We list the main notations in the paper in Table \ref{table:notations}.

\begin{table}[ht!]
\centering
\caption{Summary of notations in this paper.}
\vspace{-3mm}
\resizebox{0.49\textwidth}{!}{
\begin{tabular}{c p{8.6cm}}
\toprule
\textbf{Notation} & \textbf{Description} \\
\midrule

\( G = (V, E) \) & A graph where \( V \) is the set of nodes and \( E \) is the set of edges. \\

\( N \) & Total number of nodes in the graph. \\

\( \mathbf{A}( \hat{\mathbf{A}}) \) & The adjacency matrix of the graph and its normalized version. \\

\( \hat{\mathbf{L}} \) & Normalized Laplacian of the graph. \\

\( \mathbf{P} \) & Refers to either \( \hat{\mathbf{A}} \) or \( \hat{\mathbf{L}} \). \\

% \( \mathbf{I} \) & Identity matrix. \\

% \( \hat{\mathbf{A}} \) & Normalized adjacency matrix. \\

% \( \mathbf{D} \) & Diagonal degree matrix. \\

% \( \mathbf{U} \) & Matrix consisting of the corresponding eigenvectors of \( \mathbf{\Lambda} \). \\

% \( \mathbf{\Lambda} \) & A diagonal matrix composed of eigenvalues. \\

% \( h(\mathbf{\Lambda}) \) & Represents the graph filter operated on the diagonal matrix $\mathbf{\Lambda}$ composed of eigenvalues. \\

% \( h(\mathbf{\Lambda}) \) & Denotes the graph filter on the eigenvalue diagonal matrix $\mathbf{\Lambda}$. \\

\midrule

\( \mathbf{X} \in \mathbb{R}^{N \times d}\) & Original graph signal matrix or node feature matrix. \\

\( \mathbf{Z} \in \mathbb{R}^{N \times d} (\mathbb{R}^{N \times c} \)) & Filtered signal or representation of nodes. \\

% \( \mathbf{Z} \in \mathbb{R}^{N \times d} \text{ or } \mathbf{Z} \in \mathbb{R}^{N \times c} \) & Filtered signal or representation of nodes. \\

\midrule

\( \{g_k(\cdot)\}_{k=0}^{K} \) & Series polynomial basis of truncated order \( K \). \\

\( \{\alpha_k\}_{k=0}^{K} \) & Polynomial coefficients for all nodes, i.e.$\mathbf{Z} \approx \sum_{k=0}^{K} \alpha_k g_k(\mathbf{P})\mathbf{X}.$ \\
% $h(\lambda) \approx \sum_{k=0}^{K}\alpha_k g_k (\lambda).$ \\

\( \{\alpha_k^{(i)}\}_{k=0}^{K} \) & Polynomial coefficients of nodes $v_i$, i.e.$ \mathbf{Z}_{i,:} \approx \sum_{k=0}^{K} \alpha_k^{(i)} \left(g_k(\mathbf{P})\mathbf{X}\right)_{i,:}.$ \\
% h(\lambda) \approx \sum_{k=0}^{K}\alpha_k g_k (\lambda)

\( \{\alpha_{(p,q)k}\}_{k=0}^{K} \) & Coefficients on channel (p,q),i.e.$ \mathbf{Z}_{:,p:q} \approx \sum_{k=0}^{K} \alpha_{(p,q)k} \left(g_{k}(\mathbf{P})\mathbf{X}\right)_{:,p:q}.$ \\
% h(\lambda) \approx \sum_{k=0}^{K}\alpha_k g_k (\lambda)

% \( \{\alpha_{(p,q)k}\}_{k=0}^{K} \)

\midrule

\( \boldsymbol{h}^{(i)}_{k} \in \mathbb{R}^{d} \) & Polynomial token of order $k$ for node \( v_i \). \\

\( \mathbf{H}_k \in \mathbb{R}^{N \times d} \) & Matrix contains order-$k$ polynomial tokens for all nodes. \\

\( \mathbf{H}^{(i)}\in \mathbb{R}^{(K+1) \times d} \) & Token matrix for node \( v_i \). \\

% \( d \) & Dimension of the node representation in the graph signal matrix \( \mathbf{X} \). \\

% \( c \) & Dimension of the final node representation \( \mathbf{Z}_{i,:} \). \\

% \( h \) & Number of heads in a multi-head PolyAttn mechanism. \\

% \( d_h \) & Dimension of each channel group in a multi-head PolyAttn mechanism. \\

\midrule

% \( \boldsymbol{\beta} \in \mathbb{R}^{(K+1)} \) & Attention bias shared by all nodes. \\

% \( \mathbf{Q}, \mathbf{K}, \mathbf{V} \in \mathbb{R}^{(K+1) \times d} \) & Query, key, value matrix, respectively \\

% \( \mathbf{B} \in \mathbb{R}^{(K+1) \times d} \) & Attention bias matrix, where $\mathbf{B}_{ij} = \boldsymbol{\beta}_j$. \\

% \( \mathbf{S} \in \mathbb{R}^{(K+1) \times (K+1)} \) & Attention score matrix. \\

\( \boldsymbol{\beta} \in \mathbb{R}^{(K+1)} \) & Attention bias vector shared across all nodes. \\

\( \mathbf{B} \in \mathbb{R}^{(K+1) \times (K+1)} \) & Attention bias matrix, where each entry \( \mathbf{B}_{ij} \) equals \( \beta_j \). \\

\( \mathbf{Q}, \mathbf{K}, \mathbf{V} \in \mathbb{R}^{(K+1) \times d} \) & The query, key, and value matrices, respectively. \\

\( \mathbf{S} \in \mathbb{R}^{(K+1) \times (K+1)} \) & The attention score matrix. \\

\bottomrule
\end{tabular}
}

\label{table:notations}
\vspace{-5mm}
\end{table}

\section{Proof} \label{proof_all}
\subsection{Proof of the Theorem}

Here we provide the detailed proof for Theorem \ref{theorem1}.

\begin{proof} \label{proof_theorem}

For a node \(v_i\) in the graph, the corresponding token matrix is given by \(\mathbf{H}^{(i)} = [\boldsymbol{h}^{(i)}_0,\ldots, \boldsymbol{h}^{(i)}_K]^{\top} \in \mathbb{R}^{(K+1) \times d}\). When processed by the order-wise MLP, each row \(\mathbf{H}^{(i)}_{j,:}\) is updated as \(\mathbf{H}^{(i)}_{j,:} = \text{MLP}_{j}(\mathbf{H}^{(i)}_{j,:})\). Subsequently, the query matrix \( \mathbf{Q} \) and the key matrix \( \mathbf{K} \) are calculated as \(\mathbf{Q} = \mathbf{H}^{(i)} \mathbf{W}_Q\) and \(\mathbf{K} = \mathbf{H}^{(i)} \mathbf{W}_K\), respectively. The attention matrix \( \mathbf{A}_{attn} \in \mathbb{R}^{(K+1) \times (K+1)} \) is then formulated as $(\mathbf{A}_{attn})_{ij} = a_{ij} =(\mathbf{Q}\mathbf{K}^{\top})_{ij}$.

Taking the activation function \( \sigma \) and the attention bias matrix $\mathbf{B}$ into account, the corresponding attention score matrix $\mathbf{S} = \mathbf{A}_{attn} \odot \mathbf{B} \in \mathbb{R}^{(K+1)\times (K+1)}$, where $\mathbf{B}_{ij} = {\beta}_j, j \in \{0, \ldots, K \}$.

According to $\mathbf{H'}^{(i)} = \mathbf{S} \mathbf{V}$ and $\mathbf{V} = \mathbf{H}^{(i)}$, we have:
% \begin{align}
%     \mathbf{H'}^{(i)} &= \mathbf{S} \mathbf{H}^{(i)} =     
%     \begin{bmatrix}
%     s_{00} & s_{01} & \cdots & s_{0K} \\
%     s_{10} & s_{11} & \cdots & s_{1K} \\
%     \vdots & \vdots & \ddots & \vdots \\
%     s_{K0} & s_{K1} & \cdots & s_{KK}
%     \end{bmatrix}
%     \left[\boldsymbol{h}^{(i)}_0, \ldots, \boldsymbol{h}^{(i)}_K\right]^{\top} \\
%     &=
%     \left[\sum_{k=0}^{K} s_{0k} \boldsymbol{h}^{(i)}_k, \ldots, 
%      \sum_{k=0}^{K} s_{Kk} \boldsymbol{h}^{(i)}_k\right]^{\top},
% \end{align}
\begin{align}
    \mathbf{H'}^{(i)} &= \mathbf{S} \mathbf{H}^{(i)} =  
    \left[\sum_{k=0}^{K} s_{0k} \boldsymbol{h}^{(i)}_k, \ldots, 
     \sum_{k=0}^{K} s_{Kk} \boldsymbol{h}^{(i)}_k\right]^{\top},
\end{align}
where $\mathbf{H'}^{(i)} \in \mathbb{R}^{(K+1)\times d}$. As representation of node $v_i$ is calculated by $\mathbf{Z}_{i,:} = \sum_{k=0}^{K} \mathbf{H'}^{(i)}_{k,:}$, we have:
\begin{equation}
\begin{aligned}
    \mathbf{Z}_{i,:} 
    &= \sum_{k=0}^{K} \mathbf{H'}^{(i)}_{k,:} 
    = \sum_{k=0}^{K} s_{0k} \boldsymbol{h}^{(i)}_k + \cdots + \sum_{k=0}^{K} s_{Kk} \boldsymbol{h}^{(i)}_k \\
    &= \sum_{k=0}^{K} s_{k0} \boldsymbol{h}^{(i)}_0 + \cdots + \sum_{k=0}^{K} s_{kK} \boldsymbol{h}^{(i)}_K \\
    &= \alpha_{0}^{(i)} \boldsymbol{h}^{(i)}_0 + \cdots + \alpha_{K}^{(i)} \boldsymbol{h}^{(i)}_K \\
    &= \sum_{k=0}^{K} \alpha_{k}^{(i)} \boldsymbol{h}^{(i)}_k 
    = \sum_{k=0}^{K} \alpha_{k}^{(i)} \left(g_k\left(\mathbf{P}\right)\mathbf{X}\right)_{i,:}.
\end{aligned}
\end{equation}

Here, $\alpha_{j}^{(i)} = \sum_{k=0}^{K} {s}_{kj}, j \in \{0, \ldots, K\}$ and $\alpha_{j}^{(i)}$ is computed based on the node's token matrix \( \mathbf{H}^{(i)} \). This value serves as a \textbf{node-wise} weight for the polynomial filter and is determined by both the node features and the topology information of the node \( v_i \). Consequently, the described PolyAttn mechanism functions as a node-wise filter.
\end{proof}

\subsection{Proof of the Propositions}

In the following, we present a proof for Proposition \ref{proposition1}.

\begin{proof}
For node \(v_i\), the multi-head PolyAttn mechanism employs the sub-channel of the token matrix \( \mathbf{H}^{(i)}_{:, jd_h : (j+1)d_h - 1} \) for head \( j \), where \( j \in \{0, \ldots, h-1\} \).

According to Theorem \ref{theorem1}, there exists a set of node-wise coefficients for node \(v_i\), denoted by \( \alpha^{(i)}_{(jd_h, (j+1)d_h - 1)k} \), with \( k \in \{0, \ldots, K \} \). These coefficients are computed based on the corresponding sub-channel of the token matrix \( \mathbf{H}^{(i)}_{:,jd_h : (j+1)d_h - 1} \). The contribution of head \( j \) to the node representation \( \mathbf{Z}_{i,:} \) can then be formally expressed as:
\begin{equation}
\mathbf{Z}_{i, jd_h : (j+1)d_h - 1} = \sum_{k=0}^{K} \alpha^{(i)}_{(jd_h, (j+1)d_h - 1)k} (g_k(\mathbf{P})\mathbf{X})_{i,jd_h : (j+1)d_h - 1}.
\end{equation}

By concatenating the contributions from all heads, we obtain the complete node representation for node \(v_i\). Throughout this procedure, the multi-head PolyAttn mechanism performs a filtering operation on each channel group separately. 
\end{proof}

Below, we deliver a detailed proof for Proposition \ref{proposition0}.

\begin{proof}
According to Proof \ref{proof_theorem}, when the PolyAttn functions as a node-wise filter for node \(v_i\), we have:
\[
\mathbf{Z}_{i,:} = \sum_{k=0}^{K} \mathbf{H'}^{(i)}_{k,:} = \sum_{k=0}^{K} \alpha_k^{(i)} \left(g_k\left(\mathbf{P}\right)\mathbf{X}\right)_{i,:},
\]
where
\[
\alpha_j^{(i)} = \sum_{k=0}^{K} s_{kj} = \sum_{k=0}^{K} \sigma(a_{kj}) \beta_{j}, j \in \{0, \ldots, K\}.
\]

If the softmax function is employed, then for any node \(v_i\) in the graph, the value of \(\sum_{k=0}^{K} \sigma(a_{kj})\) remains positive after the softmax operation. The sign of \(\alpha_j^{(i)}\) is thus determined by the bias \(\beta_{j}\). Since this bias is not node-specific, it implies that the coefficients of all nodes are constrained by the bias \(\beta_{j}\), thereby limiting the expressive power of PolyAttn when acting as a node-wise filter. For instance, when all biases \(\beta_{j}\) are positive, then 
\[
\alpha_j^{(i)} = \sum_{k=0}^{K} s_{kj} = \sum_{k=0}^{K} \sigma(a_{kj}) \beta_{j} > 0,
\]
PolyAttn with a Monomial basis can only serve as a low-pass filter for all nodes \citep{chien2021GPR-GNN}. In contrast, the activation function \( \tanh(\cdot) \) allows the coefficient \(\alpha_j^{(i)} = \sum_{k=0}^{K} s_{kj} = \sum_{k=0}^{K} \sigma(a_{kj}) \beta_{j}\) to vary across nodes, enhancing the expressive power of PolyAttn.
\end{proof}

\section{Implementation Details} \label{implementation_details}

\textbf{Multi-head PolyAttn.}
Here we provide pseudocode for the multi-head PolyAttn mechanism as below.

% ---------------------------------------------------------------------------------------------------------------------
% \resizebox{0.39\textwidth}{!}{
\begin{algorithm}[t]\label{multi-head_polyattn}
\DontPrintSemicolon
\SetKwInput{Input}{Input}
\SetKwInput{Output}{Output}

\Input{Token matrix for node $v_i$: $\mathbf{H}^{(i)}=[\boldsymbol{h}^{(i)}_0,\ldots, \boldsymbol{h}^{(i)}_K]^{\top} \in \mathbb{R}^{(K+1) \times d}$}

\Output{New token matrix for node $v_i$: $\mathbf{H'}^{(i)} \in \mathbb{R}^{(K+1)\times d}$}

\Parameter{
% $\beta$, $\gamma$, $\alpha$
% \textbf{Learnable parameters:}
Projection matrix $\mathbf{W}_Q$, $\mathbf{W}_K \in \mathbb{R}^{d \times (d_h \times h)}$, \\
token-wise MLP$_j(j =0, \ldots, K)$, \\ 
attention bias $\mathbf{B}\in \mathbb{R}^{(h \times (K+1)}$ \;
}
Initialize $\mathbf{V}$ with $\mathbf{H}^{(i)}$\;

\For{j = $0$ \textnormal{\textbf{to}} $K$}{
   $\mathbf{H}^{(i)}_{j,:} \leftarrow \textnormal{MLP}_j(\mathbf{H}^{(i)}_{j,:}) $\;
}

$\mathbf{Q} \leftarrow \mathbf{H}^{(i)}  \mathbf{W}_Q$ via projection matrix $\mathbf{W}_Q$; $\mathbf{K} \leftarrow \mathbf{H}^{(i)}  \mathbf{W}_K$ via projection matrix $\mathbf{W}_K$\;

Reshape $\mathbf{Q},\mathbf{K}$ into $h$ heads to get $\mathbf{Q}_{(m)} \in \mathbb{R}^{(K+1)\times d_h}, \mathbf{K}_{(m)} \in \mathbb{R}^{(K+1)\times d_h}, m \in \{0, \ldots, h-1\}$ \\

\For{m = $0$ \textnormal{\textbf{to}} $h-1$}{
   $\mathbf{S}_{(m)} \leftarrow  \text{tanh}(\mathbf{Q}_{(m)}\mathbf{K}_{(m)}^{\top}) \odot \mathbf{B}_{m,j}$\;
   $\mathbf{H'}^{(i)}_{(m)} \leftarrow \mathbf{S}_{(m)} \mathbf{V}_{:p,q}$, where $(p,q) = (d_h \times m, d_h \times (m+1) - 1)$ \;
}

$\mathbf{H'}^{(i)} \leftarrow  [ \mathbf{H'}^{(i)}_{(0)} || \cdots || \mathbf{H'}^{(i)}_{(h-1)} ] \in \mathbb{R}^{(K+1)\times d}$, where $[\cdot||\cdots||\cdot]$ means concatenating matrices\;

\Return $\mathbf{H'}^{(i)}$ \quad  

% \# The  representation of node $v_i$ after PolyAttn is $\boldsymbol{Z}_{i,:} = \sum_{k=0}^{K} \mathbf{H'}^{(i)}_{k,:} \in \mathbb{R}^d$.

\caption{Pseudocode for Multi-head PolyAttn}
\end{algorithm}
% \vspace{-1mm}
% }
% -----------------------------------------------------------------------------------------------------

\textbf{Attention Bias.}
In implementation, we imposed constraints on the bias corresponding to each order of polynomial tokens. Specifically, for the learnable bias $\boldsymbol{\beta}$, the attention bias matrix $\mathbf{B}\in \mathbb{R}^{(K+1)\times(K+1)}$ is defined as $\mathbf{B}_{ij} = \frac{\beta_j}{ (j+1)^r}$, where hyperparameter $r$ is the constraint factor.

\textbf{Order-wise MLP.}
To enhance the expressive capacity of the order-wise MLP, we use the hyperparameter $m$ to increase the intermediate dimension of the order-wise MLP. Specifically, for an input dimension $d$ the intermediate dimension of the order-wise MLP is $m \times d$.

\section{Experimental Settings}

\subsection{PolyAttn Experiments}
\subsubsection{Fitting Signals in Synthetic Datasets} \label{exp_polyatnn_sbm}
Based on the raw signal of each node in a graph, we apply one of two predefined filters. For example, for nodes with signals $\boldsymbol{x}_1 < 0.5$, we define a low-pass filter $h_1(\lambda) = \exp(-10 \lambda^2)$, resulting in a filtered signal $\boldsymbol{z}_1=\mathbf{U}h_1(\mathbf{\Lambda})\mathbf{U}^\top\boldsymbol{x}_1$. Conversely, for nodes with signals $\boldsymbol{x}_1 \geq 0.5$, we implement a high-pass filter $h_2(\lambda) = 1 - \exp(-10 \lambda^2)$, yielding the corresponding filtered signal $\boldsymbol{z}_2=\mathbf{U}h_2(\mathbf{\Lambda})\mathbf{U}^\top\boldsymbol{x}_2$. For eigenvalues $\lambda \in [0,2]$, the predefined filters $h_1(\lambda)$ and $h_2(\lambda)$ are shown in Table \ref{predefined_filter}. Given the original graph signals $\boldsymbol{x}_1, \boldsymbol{x}_2$ and the filtered graph signals $\boldsymbol{z}_1, \boldsymbol{z}_2$, Models are expected to learn these filtering patterns.

In this experiment, every model uses a truncated order of $K=10$ within one layer. Additionally, We employed one head for PolyAttn. All models have total parameters of approximately $50,000$, achieved by using an adaptive hidden dimension.

\subsubsection{Performance on Real-World Datasets} \label{exp_polyattn_real}

To ensure a fair comparison, the truncated order \(K\) is set to \(10\) for the Monomial, Bernstein, and Chebyshev bases, and \(5\) and \(10\) for the optimal basis. The number of layers is set to \(1\) for both the node-unified filter and PolyAttn. Additionally, the number of heads for PolyAttn is one.

Hyperparameters, including hidden dimensions, learning rates, and weight-decay rates, are fine-tuned through 200 rounds of Optuna \citep{optuna_akiba2019optuna} hyperparameter search. The best configuration is chosen based on its performance on the validation set. The final outcomes are the averages of 10 evaluations on the test set with a 95\% confidence interval using the optimal parameters. 

The Optuna search space consists of 100 trials, with the searching space provided below:

% \begin{itemize}[itemsep=0pt,parsep=0pt,label=\textemdash]
% \begin{itemize}[itemsep=0pt,parsep=0pt]
\begin{itemize}[leftmargin=*,itemsep=0pt,parsep=0pt]
\item Hidden dimension: \(\{16, 32, 64, 128, 256\}\);
\item Learning rates: $\{\text{5e-5, 2e-4, 1e-3, 1e-2}\}$;
\item Weight decays: $\{\text{0.0, 1e-5, 1e-4, 5e-4, 1e-3}\}$;
\item Dropout rates:  \(\{0.0, 0.1, 0.2, \ldots, 0.9\}\);
\end{itemize}

There is one extra hyperparameter for PolyAttn:
% \begin{itemize}[itemsep=0pt,parsep=0pt]
\begin{itemize}[leftmargin=*,itemsep=0pt,parsep=0pt]
    \item Multiplying factor $m$ for order-wise MLP: \{1.0, 2.0, 0.5\}.
    % \item Dropout rate on PolyAttn: \(\{0.0, 0.1, 0.2, \ldots, 0.9\}\).
    % attn_lr [1e-2, 1e-3, 2e-4, 5e-5])
    % attn_wd [0.0, 1e-5, 1e-4, 5e-4, 1e-3])
\end{itemize}

\subsection{PolyFormer Experiments}

\subsubsection{Node Classifications} \label{exp_polyformer_common}

We train all models with the Adam optimizer \citep{adam}. Early stopping is employed with the patience of 250 epochs out of a total of 2000 epochs. The mean test accuracy, along with a 95\% confidence interval, is reported based on 10 runs.

Hyperparameter selection is carried out on the validation sets.  To expedite the hyperparameter selection process, we utilize Optuna \citep{optuna_akiba2019optuna}, performing a maximum of 400 complete trials within the following hyperparameter ranges:

% \begin{itemize}[itemsep=0pt,parsep=0pt]
\begin{itemize}[leftmargin=*,itemsep=0pt,parsep=0pt]
\item Truncated order $K$ of polynomial tokens: \( \{2, 4, 6, 8, 10, 12, 14\}\);
\item Number of layers: \(\{1, 2, 3, 4\}\);
\item Number of heads: \(\{1, 2, 4, 8, 16\}\);
\item Hidden dimension: \(\{16, 32, 64, 128, 256\}\);
\item Hidden size for FFN: \(\{32, 64, 128, 256, 512\}\);
\item Learning rates: $\{0.00005, 0.0001, 0.0005, 0.001, 0.005, 0.01\}$;
\item Weight decays: $\{\text{0.0, 1e-8, 1e-7, 1e-6, 1e-5, 1e-4, 1e-3}\}$;
\item Dropout rates: \(\{0.0, 0.1, 0.2, \ldots, 0.9\}\);
\item Constraint factor $r$: \(\{1.0, 1.2, 1.4, 1.6, 1.8, 2.0\}\);
\item Multiplying factor $m$ for order-wise MLP : \(\{1.0, 2.0, 0.5\}\).
% \item {dprate} : \(\{0.0, 0.1, 0.2, \ldots, 0.9\}\).
\end{itemize}

\subsubsection{Node Classifications on Large-Scale Datasets} \label{exp_polyformer_large}

The reported results for GNNs are sourced from \citet{chebnetii_he2022convolutional}, whereas some results for the Graph Transformer are sourced from \citet{SGFormer}. The remaining results are derived from recommended hyperparameters or through hyperparameter searching. The mean test accuracy, accompanied by a 95\% confidence interval, is reported based on either 5 or 10 runs.

We utilize the Adam optimizer \citep{adam} to train our models. Early stopping is implemented with patience at 250 epochs within an overall training span of 2000 epochs. The hyperparameter space used for experiments on large-scale datasets is enumerated below:

\begin{itemize}[leftmargin=*,itemsep=0pt,parsep=0pt]
\item Truncated order $K$ of polynomial tokens: $\{4,8,10\}$;
\item Number of layers: $\{1,2\}$;
\item Number of heads: $\{1, 4, 8\}$;
\item Hidden dimension: $\{128, 512, 1024\}$;
\item Hidden size for FFN: $\{512, 1024\}$;
\item Learning rates: $\{0.00005, 0.0002, 0.01\}$;
\item Weight decays: $\{0.0, 0.00005, 0.0005, 0.001\}$;
\item Dropout rates: $\{0.0, 0.25, 0.4, 0.5\}$;
\item Constraint factor $r$: $\{1.0, 2.0\}$;
\item Multiplying factor $m$ for order-wise MLP: $\{0.5, 1.0\}$;
\item Batch size: $\{10000, 20000, 50000\}$.
% \item \textbf{dprate}: 0.6, 0.8, 0.0
% \item \textbf{attn\_lr}: 0.01, 5e-05, 0.0002
% \item \textbf{attn\_wd}: 0.001, 0.0, 0.0005
\end{itemize}

\end{document}